\documentclass{article} 
\PassOptionsToPackage{table}{xcolor}
\usepackage{colm2024_conference}
\usepackage[T1]{fontenc}

\usepackage{amsmath,amsfonts,bm}

\def\eqref#1{equation~\ref{#1}}

\def\1{\bm{1}}

\usepackage{wrapfig}

\DeclareMathAlphabet{\mathsfit}{\encodingdefault}{\sfdefault}{m}{sl}
\SetMathAlphabet{\mathsfit}{bold}{\encodingdefault}{\sfdefault}{bx}{n}

\usepackage{amssymb}

\usepackage{hyperref}
\usepackage{url}
\usepackage{graphicx}
\graphicspath{{figure/}}
\usepackage{booktabs}
\usepackage{multirow}
\usepackage{xcolor}
\usepackage{amsmath}
\usepackage{algorithm}
\usepackage{algpseudocode}
\usepackage{fancyvrb}
\usepackage{etoc}
\newcommand{\methodname}{ExpActivator}
\newcommand{\methodnameletters}[4]{\textcolor{LeapL}{#1}\textcolor{LeapE}{#2}\textcolor{LeapA}{#3}\textcolor{LeapP}{#4}}
\newcommand{\methodnametitle}{\expandafter\methodnameletters\methodname}

\definecolor{LeapLatent}{RGB}{147,51,234}    
\definecolor{LeapExperience}{RGB}{37,99,235}
\definecolor{LeapContext}{RGB}{234,88,12}
\definecolor{LeapBehavior}{RGB}{5,150,105}

\definecolor{LeapL}{RGB}{30,64,175}   
\definecolor{LeapE}{RGB}{37,99,235}   
\definecolor{LeapA}{RGB}{59,130,246}  
\definecolor{LeapP}{RGB}{96,165,250}  

\definecolor{myblue}{RGB}{18,79,158}
\definecolor{StepBlue}{HTML}{0F4C93}   
\definecolor{TaskOrange}{HTML}{B8430D} 
\definecolor{TableHighlight}{RGB}{240,240,240}
\hypersetup{hidelinks,colorlinks=true, citecolor=myblue, urlcolor=myblue,}

\title{Relevance Does Not Imply Applicability: Experience Activation for Personal GUI Agents}

\author{Fuyao Zhang, Xuan Wang, Zherui Li, Jiaming Zhang, Longtao Huang, Wei Yang Bryan Lim \\
\vspace{5pt}\\
NTU \& Alibaba\\
}

\makeatletter
\renewenvironment{abstract}
  {\vskip.075in\centerline{\large\bf Abstract}\vspace{0.5ex}\begin{quote}}
  {\par\end{quote}\vskip 1ex}
\makeatother

\renewcommand{\weblink}{https://fyzhang1.github.io/ExpActivator}
\renewcommand{\weblinktext}{fyzhang1.github.io/ExpActivator}

\begin{document}
\etocdepthtag.toc{mainmatter}

\maketitle

\begin{abstract}
Personal Graphical User Interface (GUI) agents rely on interaction history to infer what a user wants from ambiguous instructions and to anticipate recurring routines. Existing approaches retrieve task-relevant history and append it to the policy's context, implicitly assuming that experience relevant to a task remains useful for each decision within it. We find that this help is largely spent at the first decision: retrieved history strongly improves the opening step of an episode, yet provides little sustained benefit over the remaining 90\% of steps, and offers weak guidance on whether a proactive suggestion is warranted. A relevant record may tell the agent where to begin, but not which past action applies to the current screen or whether a routine is due now. The underlying issue is that \emph{relevance does not imply applicability}: relevance is determined at the task level, whereas applicability depends on the situation at decision time.
We therefore recast personalization as \emph{experience activation} and introduce \textbf{\methodname{}}, a training-free framework that activates only the experience applicable to the current situation. During execution, \methodname{} matches each new screen to historical states in the frozen GUI backbone's latent space and supplies the corresponding action as a reference. Before execution, it activates a recurring intent only when the current time and scenario provide sufficient support, and otherwise abstains. Across four GUI backbones, \methodname{} improves within-trajectory step success by 28\% on average, achieves the best personalized execution on every backbone while using about one-fifth as many history tokens, and reaches approximately 2.3$\times$ the Matthews correlation coefficient of the strongest proactive baseline. Experience pays where it is activated, not where it is appended.
\end{abstract}

\section{Introduction}
\label{sec:introduction}

\begin{wrapfigure}[20]{r}{0.46\textwidth}
    \vspace{-16pt}
    \centering
    \includegraphics[width=\linewidth]{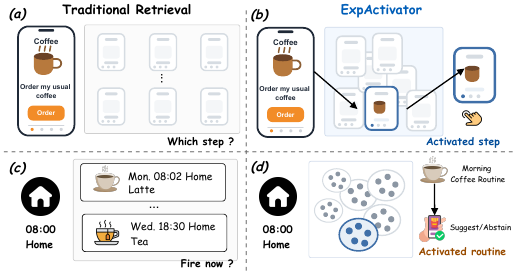}
\vspace{-6pt}
    \footnotesize
    \caption{From retrieval to activation. Top: personalized execution. Bottom:
proactive suggestion. (a)~Retrieval supplies whole trajectories without
indicating the matching step. (b)~\methodname{} activates the
matching step. (c)~Retrieval returns related records without indicating whether
a routine is due. (d)~\methodname{} activates the routine matching the
time and scenario, then suggests or abstains.}
    \label{fig:example}
\end{wrapfigure}

Personal Graphical User Interface (GUI) agents operate apps on a user's own device, where much of what the user wants is never stated explicitly~\citep{jang2026mypcbenchbenchmarkpersonallyintelligent,yuan2026osworld20benchmarkingcomputer,yang2026hippocampbenchmarkingcontextualagents,li2026weavebenchlonghorizonrealworldbenchmark}. ``Order my usual coffee'' names neither the app, the drink, nor the store; a user who orders coffee at home every morning may even expect the agent to offer it without being asked. Such gaps can only be filled from the user's interaction history~\citep{yang2025fingertip,chen2026knowubenchinteractiveproactivepersonalized,kim2026persona2webbenchmarkingpersonalizedweb}: in \emph{personalized execution}, history helps resolve an ambiguous instruction and carry it out the way this user would; in \emph{proactive suggestion}, it helps decide whether a recurring routine is due before any instruction is given.\footnote{\small{Throughout the paper, a \emph{task} is a user goal such as ordering coffee, an \emph{episode} is one execution of a task on the device, a \emph{step} is a single screen within an episode together with the action taken on it, and a \emph{routine} is a task that the user repeats at similar times and in similar scenarios, such as at home or while commuting.}} The question is therefore not whether history helps, but when and how past experience should influence each decision.

Existing personalized GUI agents primarily use history through task-level retrieval: they retrieve related records, profiles, or intent memories and append them to the policy's context~\citep{chai2026pira,yang2025fingertip,kong2026proactivemobile,lyu2026personalalign}. This implicitly treats task-level relevance as a sufficient proxy for decision-level applicability: once experience is retrieved for a task, the same context is expected to remain useful as subsequent decisions change. For execution, this leaves a sharper question: not only how much retrieved history helps, but for how long.

Our experiments reveal a sharp limitation of task-level retrieval. In execution,
retrieved history strongly improves the opening decision, which commits the
agent to an application, yet provides little sustained benefit over the
remaining 90\% of steps. This is because the instruction stays fixed while the
screen changes at every step. A retrieved trajectory may indicate where to
begin, but it does not tell the agent which past action fits \emph{this} screen
(Figure~\ref{fig:example}a). Proactive suggestion faces a similar problem. The
same routine may be retrieved both at 8~a.m.\ at home and during an evening
commute, even though it is appropriate only in the former~\citep{chen2026knowubenchinteractiveproactivepersonalized,wang2026me,nathani2026proactive,liu2026atmem,lin2026mobileguiagentprivacy}.
Even methods that organize routines by time and scenario~\citep{lyu2026personalalign}
still leave the policy to decide whether the routine applies. Both cases stem
from a gap between relevance and applicability. Retrieval identifies experience
related to the task, whereas whether that experience applies depends on the
situation at decision time.

We therefore recast personalization as \emph{experience activation}: retrieval narrows history to relevant candidates, while the situation at each decision activates the experience that applies. We instantiate this principle in \textbf{\methodname{}}, a training-free framework operating at two granularities. During execution, the current screen activates the matching historical state in the frozen GUI backbone's latent space, and its associated action is supplied as a reference; the match is recomputed as the interface changes. Before execution, the current time and scenario determine whether a recurring intent is sufficiently supported, and the agent abstains otherwise~(Figure~\ref{fig:example}b,d). Personalization thus becomes a decision-time selection problem rather than a one-time retrieval problem.

Across four GUI backbones, \methodname{} improves within-trajectory step success
by 28\% on average, achieves about 20\% higher overall step success than the
strongest baseline while using about one-fifth as many history tokens, and
reaches approximately 2.3$\times$ its Matthews correlation coefficient for
proactive suggestion. Controlled comparisons show that the same history helps
substantially more when its applicable action is selected than when it is
appended in full. Experience pays where it is activated, not where it is
appended. Our contributions are as follows:
\begin{itemize}
    \item \textbf{A diagnosis of task-level retrieval.} We show that task-level
    relevance does not imply decision-level applicability. Retrieved history
    helps mainly at the opening step of an episode and offers weak guidance on
    when a routine is due.
    \item \textbf{Experience activation at two levels.} We introduce
    \methodname{}, a training-free framework with step-level and task-level
    activation. Step-level activation matches every new screen to a historical
    state and uses its action as a reference, while task-level activation
    suggests a recurring intent only when the current time and scenario support
    it.
    \item \textbf{Evidence for decision-time activation.} Across four GUI
    backbones, step-level activation consistently improves personalized
    execution and task-level activation improves proactive suggestion, while
    using substantially less history context than existing approaches.
\end{itemize}



\begin{figure}[ht]
  \centering
  \includegraphics[width=\linewidth]{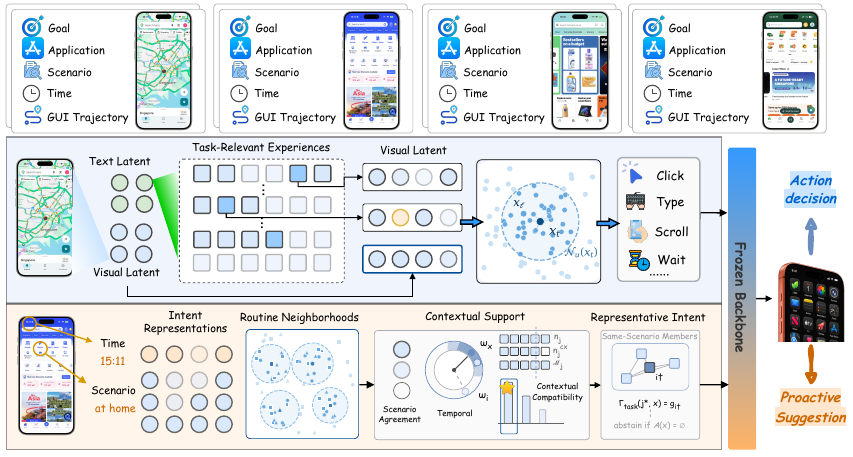}
  \caption{Overview of \methodname{}. \textcolor{StepBlue}{Step-level activation}
  (middle) matches each screen to recorded states of task-relevant trajectories
  and supplies the best-matching action as a reference.
  \textcolor{TaskOrange}{Task-level activation} (bottom) scores recurring intents
  by scenario and temporal support and supplies a representative intent, or
  abstains. A frozen backbone outputs an action or suggestion.}
  \label{fig:framework}
  \vspace{-5pt}
\end{figure}

\section{Method}
\label{sec:method}

\subsection{Problem Formulation}
\label{sec:problem-definition}

For user $u$, let $\mathcal M_u=\{r_i\}_{i\in\mathcal I_u}$ denote the
interaction history. Each record $r_i=(g_i,p_i,c_i,\tau_i,\xi_i)$ contains
a goal, application, scenario, timestamp, and trajectory.
The trajectory $\xi_i=\{(s_{i,j},a_{i,j})\}_{j=0}^{T_i-1}$ pairs each
screen with the action taken from it. We consider two settings:
instruction-driven execution, which predicts an action given a goal
$q$, current screen $s_t$, and preceding actions $\mathcal H_t$; and
proactive suggestion, which proposes a routine or abstains given a
scenario and timestamp $(c_x,\tau_x)$. \methodname{} addresses both settings through \emph{experience activation}:
selecting historical behavior for the current decision rather than
reusing an entire experience as a fixed script. Activation operates at
two granularities. Step-level activation selects a historical action
for execution, whereas task-level activation selects a historical
intent for proactive suggestion. For $d\in\{\mathrm{step},\mathrm{task}\}$,
let $x=(q,s_t)$ in the former case and $x=(c_x,\tau_x)$ in the latter.
For a nonempty candidate set, we define
\begin{equation}
    e_u^{(d)}(x)
    =
    {\Gamma_d}\!\Bigl(
        \operatorname*{arg\,max}_{
            b\in{\color{StepBlue}\mathcal B_u^{(d)}(x)}
        }
        \;{\color{TaskOrange}\kappa_d(x,b)},
        \;x
    \Bigr).
    \label{eq:experience-activation}
\end{equation}
Here, $\mathcal B_u^{(d)}(x)$ indexes the candidate historical steps or
routines, $\kappa_d$ measures their compatibility with the current
input, and $\Gamma_d$ returns the associated action or intent.
Frozen latent representations establish the semantic and visual
relationships used for activation; time and scenario remain explicit
contextual variables. Figure~\ref{fig:framework} traces both
instantiations, which Sections~\ref{sec:step-activation}
and~\ref{sec:task-activation} define in turn.

\subsection{Step-level Activation}
\label{sec:step-activation}

Given $q$, \methodname{} first restricts retrieval to task-relevant experiences
(Figure~\ref{fig:framework}, middle).
With a unit-normalized text encoder $f$, record relevance is
$\alpha_i(q)=f(q)^\top f(\rho(g_i))$, where $\rho$ removes the leading
application phrase while retaining task content. We select
$\mathcal C_u(q)=
\operatorname{TopK}_{i\in\mathcal I_u^\xi;\,K_\xi}\alpha_i(q)$,
where $\mathcal I_u^\xi$ contains records with readable trajectories.
Before the first observation the same records also supply an initial
reference $e_0$; its construction is routine. For subsequent observations $x_t=(q,s_t)$, the candidates are
$\mathcal B_u^{(\mathrm{step})}(x_t)
=\{(i,j):i\in\mathcal C_u(q),\,0\leq j<T_i\}$.
To compare the current screen with these historical states, \methodname{} uses
the GUI backbone's visual features. Let $v_k(s)$ be a unit-normalized
pooled token block and $\mu_k$ its mean over historical calibration
images. The screen representation is
\begin{equation}
    \phi(s)=\frac{1}{\sqrt B}
    \left[
        \nu\!\left(v_1(s)-\mu_1\right);
        \ldots;
        \nu\!\left(v_B(s)-\mu_B\right)
    \right],
    \label{eq:visual-state-representation}
\end{equation}
where $\nu$ denotes $\ell_2$ normalization. The $B$ blocks follow a
spatial layout, optionally augmented with a global block, selected
through historical calibration for each backbone. The compatibility
score
$\kappa_{\mathrm{step}}(x_t,(i,j))
=\phi(s_t)^\top\phi(s_{i,j})$
therefore averages block-wise cosine similarities between centered
features. Equation~\ref{eq:experience-activation} selects a historical step
jointly over trajectories and their local states:
\begin{equation}
    (i_t^*,j_t^*)=
    \arg\max_{\substack{i\in\mathcal C_u(q)\\0\leq j<T_i}}
    \phi(s_t)^\top\phi(s_{i,j}),
    \qquad
    e_t=\Gamma_{\mathrm{step}}\bigl((i_t^*,j_t^*),x_t\bigr).
    \label{eq:step-activation}
\end{equation}
The operator $\Gamma_{\mathrm{step}}$ serializes the recorded action
$a_{i_t^*,j_t^*}$ into the policy's reference format.
While $\mathcal C_u(q)$ remains fixed within a task, the selected step
is recomputed after every observation. The reference can therefore
move between historical steps and trajectories without assuming that
the current execution follows their step indices. The frozen policy generates
$\widehat a_t=\pi_\theta(q,s_t,\mathcal H_t,e_t)$.
The historical action serves as a reference, not a prescribed operation:
the policy also observes the current screen to assess applicability
and locate the target. State matching and policy inference repeat as
execution progresses, allowing the referenced experience to change
with the observed interface.

\subsection{Task-level Activation}
\label{sec:task-activation}

Without an explicit goal, \methodname{} identifies repeated tasks
(Figure~\ref{fig:framework}, bottom) from complete
historical intents. Let $z_i=\psi(g_i)$ be the unit-normalized embedding
from an intent encoder. Each center $j\in\mathcal I_u$ defines a routine
neighborhood
$\mathcal R_j(x)=
\{i\in\mathcal I_u:\tau_i\neq\tau_x,\ z_j^\top z_i\geq\eta\}$,
excluding records at the query timestamp. These neighborhoods group
semantically similar occurrences and may overlap. A routine's support under the current context combines scenario
agreement with temporal proximity:
\begin{equation}
    M_j(x)=
    \sum_{i\in\mathcal R_j(x)}
    \mathbb I[c_i=c_x]\,
    \exp\!\left(
        -\frac{d_{24}(\omega_i,\omega_x)^2}{2\sigma^2}
    \right),
    \label{eq:contextual-mass}
\end{equation}
where $\omega_i=\operatorname{hour}(\tau_i)\in[0,24)$ and
$d_{24}(a,b)=\min(|a-b|,24-|a-b|)$.
Let $n_j=|\mathcal R_j(x)|$ denote total support and
$n_j^{c_x}=\sum_{i\in\mathcal R_j(x)}\mathbb I[c_i=c_x]$
denote same-scenario support. \methodname{} distinguishes whether a routine has sufficient support for
activation from how well it matches the present context:
\begin{align}
    \mathcal A(x)
    &=
    \left\{
        j\in\mathcal I_u:
        n_j\geq n_{\min},\
        n_j^{c_x}\geq n_{\mathrm{ctx}},\
        M_j(x)\geq\delta
    \right\},
    \label{eq:activation}\\
    \kappa_{\mathrm{task}}(x,j)
    &=
    \frac{M_j(x)}{n_j^\gamma},
    \qquad \gamma=1.
    \label{eq:contextual-score}
\end{align}
Thus, absolute support determines eligibility, while mean contextual
match per occurrence determines ranking. Setting
$\mathcal B_u^{(\mathrm{task})}(x)=\mathcal A(x)$
instantiates the candidate set in
Equation~\ref{eq:experience-activation}.
If this set is empty, \methodname{} abstains; otherwise, it selects
$j^*=\arg\max_{j\in\mathcal A(x)}\kappa_{\mathrm{task}}(x,j)$. The activated intent is drawn from the selected routine's same-scenario
members,
$\mathcal R_{j^*}^{c_x}
=\{i\in\mathcal R_{j^*}(x):c_i=c_x\}$.
$\Gamma_{\mathrm{task}}(j^*,x)=g_{i^\dagger}$, where
$i^\dagger=
\arg\max_{i\in\mathcal R_{j^*}^{c_x}}
\sum_{k\in\mathcal R_{j^*}(x)}z_i^\top z_k$
is the same-scenario member most similar, on average, to the entire routine. The corresponding historical intent is
provided to the frozen backbone as a reference for suggestion
generation, not as authorization to execute the task. All encoders and policy models remain frozen. Retrieval configurations
and visual layouts are calibrated on held-out historical records and
fixed during evaluation. Encoder specifications, calibration protocols,
activation thresholds, and reference formats are detailed in
Appendix~\ref{app:cue-details}.

\section{Experiment}

\subsection{Experimental Setup}
\label{sec:setup}

\paragraph{Dataset.}
We evaluate on AndroidIntent~\citep{lyu2026personalalign}, built from long-term
mobile records of 82 users across 135 applications and 13 scenario categories.
The chronological first 80\% of each user's records forms the history
$\mathcal M_u$ (15,966 records), and the remaining episodes are evaluation
targets excluded from retrieval. Personalized execution has 775 episodes with
7,915 steps whose instructions are deliberately ambiguous: a request such as
checking updates names neither application nor content, so the intended
behavior must be recovered from history. Proactive suggestion has 315 states
(215 positive, 100 negative), each giving only a user, a time, and a scenario.

\paragraph{Metrics.}
For execution we report action-type accuracy (Type), step success rate (SSR),
and the benchmark's cumulative success rate (CSR). For proactive suggestion we report the official
semantic score of emitted suggestions (Sem.), precision (P), balanced accuracy
(BA), and the Matthews correlation coefficient (MCC), which is zero for any rule
that cannot tell positive from negative states; recall, false-alarm rate, and
F1 are listed in Appendix~\ref{app:proactive-complete}. Comparisons use paired
bootstrap resampling, and all configurations are calibrated on held-out history
before evaluation (Appendix~\ref{app:cue-details}).

\paragraph{Backbones and baselines.}
We use four frozen open GUI backbones with different visual encoders and GUI
training, Qwen3-VL-8B, MAI-UI-8B, Qwen3.5-9B, and GUI-Owl-1.5-8B, with the
proprietary QwenVL-Max as an external reference. Within each backbone we compare
\methodname{} with the history mechanisms of the benchmark: \textbf{Recent}
supplies the latest records, \textbf{Retrieve} the records most similar to the
instruction, \textbf{LLM-UM} an offline LLM-written profile, and
\textbf{HIM-Agent} a hierarchical intent memory~\citep{lyu2026personalalign};
\textbf{Base} uses no history. All methods read the same history and share model
revisions and inference settings, so comparisons are paired step by step.

\begin{table*}[t]
    \centering
    \small
    \caption{Personalized execution and proactive suggestion on AndroidIntent.
    Within each block all methods share the same history and frozen policy. QwenVL-Max's BA and MCC are derived from its published recall and false-alarm rate, and Always suggest suggests on every state.
    Values are percentages, and bold marks the best result within each block.}
    \label{tab:combined-personalized-proactive}
    \resizebox{\textwidth}{!}{%
    \begin{tabular}{llccc|cccc}
      \toprule
      & & \multicolumn{3}{c|}{Personalized}
          & \multicolumn{4}{c}{Proactive} \\
      \cmidrule(lr){3-5}\cmidrule(lr){6-9}
      Backbone & Method
        & Type$\uparrow$ & SSR$\uparrow$ & CSR$\uparrow$
        & Sem.$\uparrow$ & P$\uparrow$ & BA$\uparrow$ & MCC$\uparrow$ \\
      \midrule
      \multirow{6}{*}{Qwen3-VL-8B}
        & Base & 47.6 & 19.4 & 23.7 & -- & -- & -- & -- \\
        & Recent & 50.6 & 20.6 & 29.2 & 48.6 & 68.4 & 50.3 & 2.3 \\
        & Retrieve & 51.6 & 22.6 & 41.4 & 48.4 & 67.8 & 48.9 & $-5.5$ \\
        & LLM-UM & 51.0 & 22.0 & 39.2 & 29.1 & 68.9 & 51.5 & 10.5 \\
        & HIM-Agent & 51.3 & 22.6 & 40.2 & 50.8 & 72.7 & 58.5 & 20.7 \\
      \rowcolor{TableHighlight}
        & \textbf{\methodname{}}
          & \textbf{58.1} & \textbf{30.9} & \textbf{47.9}
          & \textbf{52.1} & \textbf{82.6} & \textbf{71.8} & \textbf{42.7} \\
      \midrule
      \multirow{6}{*}{MAI-UI-8B}
        & Base & 46.6 & 17.9 & 22.3 & -- & -- & -- & -- \\
        & Recent & 47.6 & 18.1 & 21.2 & 44.5 & 68.3 & 50.0 & 0.0 \\
        & Retrieve & 49.7 & 21.6 & 32.7 & 44.3 & 68.3 & 50.0 & 0.0 \\
        & LLM-UM & 48.1 & 19.0 & 28.3 & 26.7 & 68.3 & 50.0 & 0.0 \\
        & HIM-Agent & 48.2 & 19.9 & 28.7 & \textbf{50.5} & 70.6 & 54.8 & 14.1 \\
      \rowcolor{TableHighlight}
        & \textbf{\methodname{}}
          & \textbf{56.5} & \textbf{26.6} & \textbf{42.7}
          & 49.6 & \textbf{82.7} & \textbf{72.7} & \textbf{45.2} \\
      \midrule
      \multirow{6}{*}{Qwen3.5-9B}
        & Base & 50.1 & 17.2 & 21.4 & -- & -- & -- & -- \\
        & Recent & 49.4 & 17.9 & 25.5 & 47.3 & 69.5 & 52.7 & 10.2 \\
        & Retrieve & 49.2 & 19.9 & 32.5 & \textbf{52.1} & 68.2 & 50.0 & 0.0 \\
        & LLM-UM & 52.4 & 20.7 & 39.8 & 21.5 & 61.9 & 49.0 & $-3.6$ \\
        & HIM-Agent & 51.4 & 21.0 & 37.8 & 48.4 & 82.4 & 61.2 & 22.0 \\
      \rowcolor{TableHighlight}
        & \textbf{\methodname{}}
          & \textbf{55.9} & \textbf{24.3} & \textbf{40.9}
          & 51.0 & \textbf{86.5} & \textbf{70.9} & \textbf{38.9} \\
      \midrule
      \multirow{6}{*}{GUI-Owl-1.5-8B}
        & Base & 58.2 & 22.7 & 21.5 & -- & -- & -- & -- \\
        & Recent & 59.1 & 23.9 & 27.2 & 46.5 & 69.2 & 52.1 & 9.1 \\
        & Retrieve & 60.3 & 28.2 & 41.6 & 48.2 & 67.8 & 49.1 & $-2.8$ \\
        & LLM-UM & 57.9 & 24.8 & 30.8 & 26.6 & 67.1 & 48.1 & $-3.9$ \\
        & HIM-Agent & 58.7 & 27.2 & 35.6 & 51.1 & 72.5 & 58.0 & 19.0 \\
      \rowcolor{TableHighlight}
        & \textbf{\methodname{}}
          & \textbf{61.8} & \textbf{28.9} & \textbf{45.8}
          & \textbf{53.9} & \textbf{82.6} & \textbf{72.4} & \textbf{44.6} \\
      \midrule
      QwenVL-Max & -- & 51.6 & 24.8 & 27.3 & 52.2 & 67.4 & 49.6 & $-2.4$ \\
      \multicolumn{2}{l}{Always suggest} & -- & -- & -- & -- & 68.3 & 50.0 & 0.0 \\
      \bottomrule
    \end{tabular}%
    }
  \end{table*}

\subsection{Main Results}

\begin{table}[t]
\centering\small
\caption{History-token cost and the two phases of an episode on Qwen3-VL-8B.
History tokens are prompt tokens added over the no-history condition. Opening is
each episode's first action and within-trajectory covers the 7,140 actions after
it. The last column is the within-trajectory difference from Base with paired
95\% user-cluster intervals. Rates are percentages.}
\label{tab:ExpActivator-cost-decomposition}
\resizebox{\linewidth}{!}{%
\begin{tabular}{lrrrrrl}
\toprule
& \multicolumn{2}{c}{Prompt tokens/step}
& \multicolumn{3}{c}{Step success rate}
& Within-trajectory \\
\cmidrule(lr){2-3}\cmidrule(lr){4-6}
Method & History & Total & All & Opening & Within-traj.
& $-$ Base [95\% CI] \\
\midrule
Base      &   0.0 & 1011.4 & 19.39 & 28.26 & 18.43 & -- \\
Recent    & 756.6 & 1768.0 & 20.57 & 40.90 & 18.36 & $-0.07$ [$-0.87$, $+0.85$] \\
Retrieve  & 730.6 & 1742.0 & 22.63 & 62.06 & 18.35 & $-0.08$ [$-1.16$, $+0.99$] \\
LLM-UM    & 843.7 & 1855.1 & 22.01 & 62.32 & 17.63 & $-0.80$ [$-1.83$, $+0.12$] \\
HIM-Agent & 417.8 & 1429.2 & 22.58 & 62.32 & 18.26 & $-0.17$ [$-0.89$, $+0.57$] \\
\rowcolor{TableHighlight}
\textbf{\methodname{} (ours)}
          & \textbf{144.6} & \textbf{1156.0} & \textbf{30.92}
          & \textbf{70.06} & \textbf{26.67}
          & $\mathbf{+8.24}$ [$+6.51$, $+10.16$] \\
\bottomrule
\end{tabular}}
\end{table}

\begin{figure}[t]
\centering
\includegraphics[width=\linewidth]{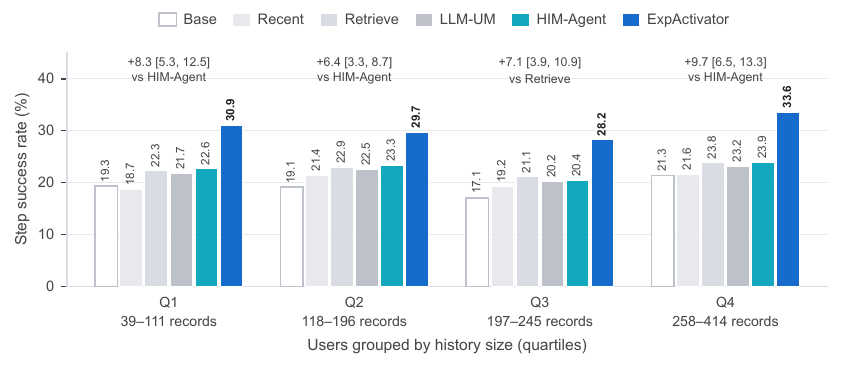}
\vspace{-5pt}
\caption{Step success by user history size on Qwen3-VL-8B. Users are grouped
into quartiles by the records in their memory split, and all six conditions are
scored on the same steps. Text above each group gives \methodname{} minus the
strongest baseline there, with paired 95\% user-cluster intervals.}
\label{fig:history-size-main}
\vspace{-10pt}
\end{figure}

\paragraph{Personalized execution.}
Table~\ref{tab:combined-personalized-proactive} shows that \methodname{} achieves
the best Type, SSR, and CSR on all four backbones, even though these backbones
differ in visual encoder and GUI training. Relative to the same model without
history, it improves SSR by 44\% on average and roughly doubles CSR, and each of
these gains is statistically significant under paired bootstrap. The margin over
the memory baselines is also substantial: \methodname{} improves SSR by about
20\% on average over the strongest baseline in each block and by 37\% on
Qwen3-VL-8B, although every baseline reads exactly the same history. With
\methodname{}, every open backbone surpasses the proprietary QwenVL-Max in CSR,
reaching about 1.6 times its score on average. The improvement therefore comes
neither from a larger policy nor from more history, but from supplying the part
of the history that applies to the current step.

\paragraph{Proactive suggestion.}
A proactive assistant is useful only if it speaks at the right moments, and F1
cannot measure this on AndroidIntent: because 68\% of states are positive,
suggesting on every state already reaches 81.1\% F1. We therefore evaluate the
decision to suggest with BA and MCC, which credit a suggestion only when it
separates the states that call for one from those that do not. By these measures
the record- and profile-based baselines remain close to chance, and on MAI-UI
three of them behave exactly like the always-suggest rule, since retrieving
related records gives them no basis for staying silent. \methodname{} reaches
about 2.3 times the MCC of the strongest baseline on average, improves BA by 24\%
relative to the best baseline, and attains the highest precision on every
backbone, 84\% on average. Deciding whether to speak before the backbone decides
what to say is what turns accumulated routines into suggestions worth emitting.

\subsection{Where Retrieved Experience Stops Paying}
\label{sec:ExpActivator-cost-coldstart}

Because every method shares the backbone and the history, we can ask not only
how much each one helps but for how long. An episode opens with one decision
that commits it to an application and then proceeds through a trajectory of
interface states, giving 775 openings and 7,140 subsequent actions. Scoring the
two separately tests the question raised in Section~\ref{sec:introduction}: does
a retrieved record remain usable once the trajectory is under way?

\paragraph{Retrieved experience is spent at the first decision.}
History matters most at the opening decision. On Qwen3-VL-8B, the strongest
baselines raise it from 28.3\% without history to 62.3\%, and \methodname{}
raises it to 70.1\% (Table~\ref{tab:ExpActivator-cost-decomposition}). Within the
trajectory the picture reverses: over the remaining 7,140 actions, none of the
four baselines is distinguishable from the no-history condition, and all stay
within 5\% of its success rate, even though they differ in what they retrieve and
how they compress it. \methodname{} improves the same actions by 45\% relative to
the no-history condition. The pattern holds on every backbone, where no baseline
adds more than 14\% within the trajectory while \methodname{} adds 28\% on
average (Appendix~\ref{app:within-all}). Retrieval tells the agent which
experience is related, and that answer is used up by the first decision; keeping
experience useful afterwards requires selecting from it again at every
observation.

\paragraph{Lower cost, from any amount of history.}
\methodname{} obtains these gains with a much lighter prompt. It supplies a single
action reference per step and adds 144.6 history tokens, nearly five times fewer
than the average baseline and nearly three times fewer than HIM-Agent, the most
economical one, while scoring highest (Table~\ref{tab:ExpActivator-cost-decomposition};
Figure~\ref{fig:leaf-cost-pareto} in Appendix~\ref{app:leaf-cost}). No
evaluated baseline therefore matches it on accuracy and cost at once. The benefit
also does not depend on a long personal history. When users are grouped into
quartiles by the size of their history, \methodname{} outperforms the strongest
baseline in every group by more than 27\% in relative SSR, and by 37\% for the
users with the fewest records (Figure~\ref{fig:history-size-main}). A modest
number of past episodes already provides states that the current screen can
activate.

\subsection{Ablation Studies}
\label{sec:leap-ablations}

Table~\ref{tab:leap-core-ablations} isolates each design choice, using the
frozen Qwen3-VL-8B policy on the 7,140 within-trajectory actions and task-level
activation on the 315 states; Appendix~\ref{app:leap-ablations} gives the
protocols.

\begin{table}[t]
\centering\small
\caption{Ablations with paired 95\% user-cluster intervals. Step-level rows
report policy SSR with one or eight retrieved trajectories held fixed;
task-level rows report balanced accuracy at the declared operating point.}
\label{tab:leap-core-ablations}
\begin{tabular}{lrrl}
\toprule
Control & Control score & \methodname{} & Gain [95\% CI] \\
\midrule
\multicolumn{4}{l}{\textit{Step-level activation: policy SSR, $K_\xi=1$}} \\
Whole trajectory & 18.47 & 23.84 & 5.36 [3.74, 7.06] \\
\midrule
\multicolumn{4}{l}{\textit{Step-level activation: policy SSR, $K_\xi=8$}} \\
Previous screen & 22.63 & 26.67 & 4.03 [3.22, 4.84] \\
Update every four steps & 23.17 & 26.67 & 3.50 [2.59, 4.47] \\
\midrule
\multicolumn{4}{l}{\textit{Task-level activation: balanced accuracy}} \\
No time weighting & 69.33 & 72.66 & 3.34 [0.15, 6.74] \\
No scenario condition & 64.58 & 72.66 & 8.08 [2.64, 13.70] \\
\bottomrule
\end{tabular}
\end{table}

\begin{figure}[t]
\centering
\includegraphics[width=\linewidth]{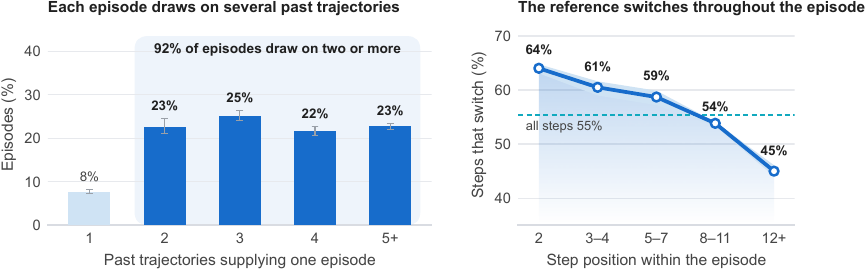}
\caption{How step-level activation draws on past trajectories. Left: number of
distinct past trajectories that supply the references of one episode. Right:
share of steps whose reference comes from a different trajectory than at the
previous step. Bars and the line give the mean over the four backbones; whiskers
and the shaded band show their range.}
\label{fig:reference-switching}
\vspace{-8pt}
\end{figure}

\paragraph{Selecting beats appending.}
To isolate the value of selection, we restrict the history to the single
top-ranked trajectory and compare two ways of using it. Supplying the
trajectory whole, as retrieval-based memories do, yields 18.5\% SSR, on par with
using no history at all. Letting step-level activation select one action from
the same trajectory raises SSR to 23.8\%, a 29\% relative improvement. The same
experience therefore helps far more when the agent receives only the part that
applies, which mirrors the within-trajectory failure of the memory baselines in
Section~\ref{sec:ExpActivator-cost-coldstart}.

\paragraph{Refresh at every observation.}
Selecting the right step also requires selecting it at the right moment. Holding
the eight retrieved trajectories fixed, we keep the matching rule but feed it a
stale screen. Matching against the previous screen lowers SSR by 15\% relative,
and refreshing the reference only every fourth step lowers it by 13\%. Agreement
between the selected reference and the recorded action drops even more sharply,
by 38\% relative, because a stale screen activates a step the user has already
passed. Recomputing the match at every observation is therefore not a
refinement of the mechanism but the mechanism itself.

\paragraph{Activation moves across trajectories.}
The refresh results imply that the activated reference should change as the
interface changes, and Figure~\ref{fig:reference-switching} shows that it does.
On every backbone, about 92\% of episodes draw their references from at least
two past trajectories, 3.4 on average, and the reference moves to a different
trajectory on 55\% of consecutive steps. Switching is most frequent early in an
episode, where generic screens resemble many past tasks, and it persists deep
into the trajectory, still occurring on nearly half of the steps after the
twelfth. These statistics are almost identical across the four backbones even
though each matches screens with its own visual features, so the behavior
follows from the activation rule rather than from any particular encoder. No
single past trajectory, however relevant, could supply these references on its
own.

\paragraph{Context gates the suggestion.}
Task-level activation relies on three contextual components, and removing each
one hurts in a distinct way. Without time weighting, BA falls by about 5\%
relative, since repetition alone does not reveal when a routine applies. Without
the scenario condition, BA falls by 11\% relative and false alarms rise from
37\% to 62\%, so scenario agreement is what keeps activation selective.
Frequency normalization leaves every trigger decision unchanged but improves the
semantic quality of emitted suggestions by 4\% relative, by favoring routines
that match the context on each occurrence rather than on a few. Time decides
when to fire, scenario decides whether to fire, and frequency decides what to
say.

\subsection{Case Study}
\label{sec:case-study}

\begin{figure}[t]
\centering
\resizebox{\linewidth}{!}{%
\includegraphics[height=4cm]{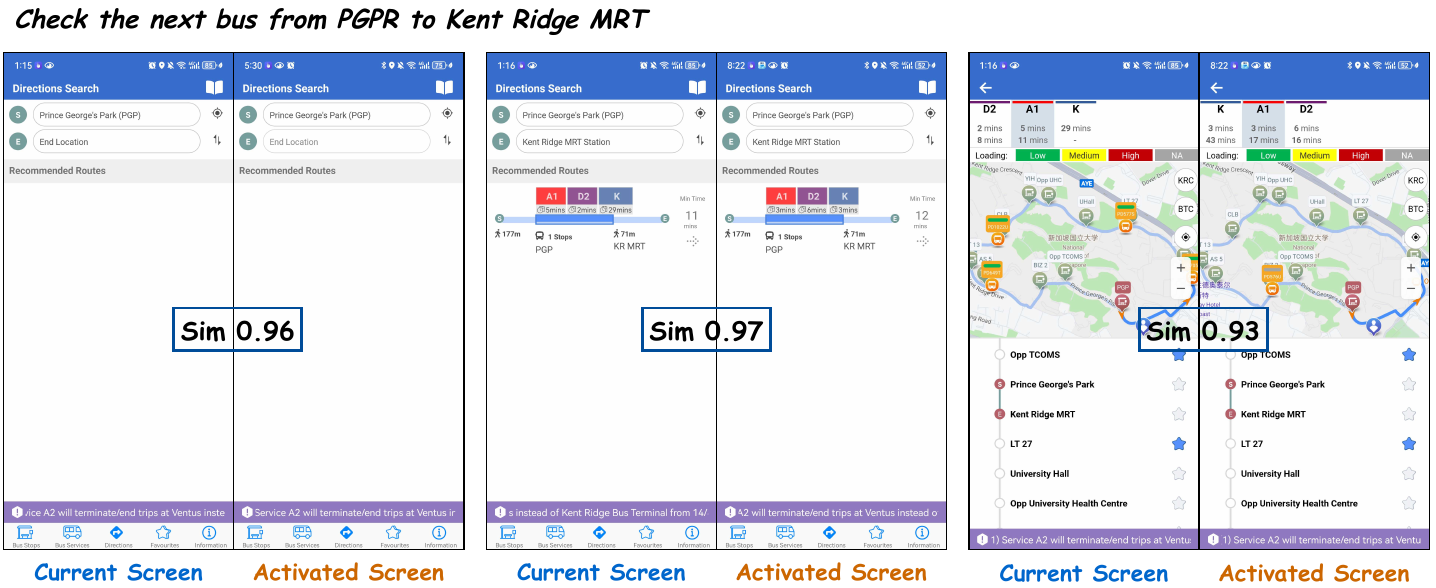}\hspace{0.35cm}%
\includegraphics[height=4cm]{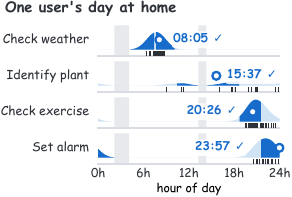}}
\caption{Activation on real records. Left: step-level activation places each
current screen of one NUS NextBus task beside the recorded state it activates.
Right: task-level activation for one user at home, showing four routines; each
curve is a routine's contextual support over the day, dark segments mark when it
is activated, gray bands mark abstention, and ticks mark its past occurrences.
Intents are translated from Chinese.}
\label{fig:task-case}
\vspace{-10pt}
\end{figure}

Figure~\ref{fig:task-case} shows both granularities of activation on real
records. On the left, each current screen of a NUS NextBus task activates a
recorded state that nearly duplicates it, and the reference moves from one past
trip to another as the screens change, following the trip that matches the
current stage of the task (Appendix~\ref{app:case-study}). On the right, the
routine with the strongest contextual support for one user changes over the
day, so task-level activation offers the weather at 08:05, plant identification
at 15:37, the day's exercise at 20:26, and an alarm at 23:57, each matching what
the user then did. Before dawn and shortly after 1 p.m., no routine has enough
support and \methodname{} stays silent. Figures~\ref{fig:task-more-a}
and~\ref{fig:task-more-b} show the same behavior for five more users.

\section{Related Work}
\label{sec:related-work}

\paragraph{Personalized GUI Agents.}
Personalized GUI benchmarks evaluate whether agents follow user
preferences across mobile tasks, with settings ranging from recorded
demonstrations to interactive preference
elicitation~\citep{yang2025fingertip,nie2026pspa,chen2026knowubenchinteractiveproactivepersonalized,li2025coloragentbuildingrobustpersonalized}.
Methods use preference learning and hierarchical memory to recover
information omitted from instructions~\citep{wang2026me,lyu2026personalalign}.
The closest precedent, PersonalAlign, introduces AndroidIntent and
HIM-Agent, which aggregates long-term records into preference and
routine prototypes~\citep{lyu2026personalalign}.
\methodname{} extends this focus on
recovering user intent to selecting the historical behavior that can
guide each decision. It distinguishes retrieving a task-relevant
trajectory from identifying the historical step appropriate for the
current screen.

\paragraph{Memory and Experience Reuse in GUI Agents.}
Reusable workflows and structured memories provide experience for GUI
planning and action
generation~\citep{wang2024awm,zhu2026hymem,qin2026eam,chung2025evaluatinglongcontextreasoningllmbased,chen2025graph2eval}.
Memory can also be refreshed or selectively managed during
execution~\citep{zhu2026hymem,zeng2026mementogui,liu2026atmem}.
MementoGUI learns multimodal memory selection and compression with a
frozen GUI backbone~\citep{zeng2026mementogui}, while ATMem tracks the
role and status of task information and learns selective memory
use~\citep{liu2026atmem}. \methodname{} contributes a mechanism for
selecting concrete actions from personal history without training a
memory controller or the GUI policy. Step-level activation matches each
screen against historical states within task-relevant trajectories and
updates the action reference after every observation. The frozen
policy combines this reference with the current screen and execution
history to produce the action. Controlled comparisons with fixed
candidate history support the value of screen matching and reference
updates for personalized action prediction.

\paragraph{Proactive Assistance.}
Proactive mobile benchmarks study how user history, screen activity,
and device context support intent inference without explicit
requests~\citep{yang2025fingertip,chai2026pira,kong2026proactivemobile}.
Interactive evaluations further examine intervention timing and
responses to feedback~\citep{chen2026knowubenchinteractiveproactivepersonalized,nathani2026proactive}.
Closest to our setting, HIM-Agent uses temporal and scenario
regularities to organize routine memory and supplies representative
intents and states for proactive reasoning~\citep{lyu2026personalalign}.
\methodname{} uses this contextual information in an explicit
activation rule applied before suggestion generation. Task-level
activation separates the support required to consider a routine from
the contextual score used to rank it, and abstains when no routine
qualifies. \methodname{} provides a common basis for using personal
history across execution and suggestion: current conditions select
an action or intent reference, while the frozen backbone generates
the response. Both thus adapt their use of experience without
fine-tuning.
\section{Conclusion}

We have argued that personal experience helps only where it is activated:
a relevant trajectory need not contain an action suited to the current
screen, and a familiar routine may become due even without any instruction. 
\methodname{} acts on this principle without any training. Step-level
activation re-anchors the action reference to every observation, while
task-level activation fires only when time and scenario support a
routine. The same record store then yields, across four frozen backbones,
the best personalized execution and the only selective proactive
suggestion among the evaluated methods, at the lowest history-token cost.
For personal agents, the lesson is compact: experience pays where it is
activated, not where it is appended.

\bibliography{colm2024_conference}
\bibliographystyle{colm2024_conference}

\clearpage
\newpage
\appendix

\etocdepthtag.toc{appendix}
\begingroup
\hypersetup{linkcolor=myblue}
\etocsettagdepth{mainmatter}{none}
\etocsettagdepth{appendix}{subsection}
\etocsettocstyle{\section*{Appendix}}{}
\tableofcontents
\clearpage

\section{Notation and Algorithms}
\label{app:algorithms-section}

\subsection{Notation}
\label{app:notation}

Table~\ref{tab:notation} collects the symbols used in Section~\ref{sec:method}.

\begin{table}[htbp]
\centering
\small
\caption{Notation.}
\label{tab:notation}
\begin{tabular}{@{}ll@{}}
\toprule
Symbol & Meaning \\
\midrule
$\mathcal M_u,\ \mathcal I_u$ & interaction history of user $u$ and its record indices \\
$r_i=(g_i,p_i,c_i,\tau_i,\xi_i)$ & record: goal, application, scenario, timestamp, trajectory \\
$\xi_i=\{(s_{i,j},a_{i,j})\}_{j<T_i}$ & recorded screens and the actions taken from them \\
$q,\ s_t,\ \mathcal H_t$ & instruction, current screen, preceding actions \\
$(c_x,\tau_x)$ & scenario and timestamp of a proactive query \\
$\mathcal B_u^{(d)}(x),\ \kappa_d,\ \Gamma_d$ & candidates, compatibility, and readout at granularity $d$ \\
$f,\ \alpha_i(q),\ K_\xi,\ \mathcal C_u(q)$ & goal encoder, record relevance, number and set of candidate trajectories \\
$\phi(s),\ B,\ \mu_k$ & centered block screen representation, number of blocks, block means \\
$(i_t^*,j_t^*),\ e_t$ & activated historical step and its action reference at step $t$ \\
$\psi,\ z_i,\ \eta$ & intent encoder, intent embedding, routine threshold \\
$\mathcal R_j(x),\ n_j,\ n_j^{c_x}$ & routine neighborhood, total and same-scenario support \\
$M_j(x),\ \sigma$ & contextual mass and temporal bandwidth \\
$n_{\min},\ n_{\mathrm{ctx}},\ \delta,\ \gamma$ & support, scenario, mass thresholds and frequency exponent \\
$j^*,\ i^\dagger$ & activated routine and its representative intent \\
\bottomrule
\end{tabular}
\end{table}

\subsection{Algorithms}
\label{app:algorithms}

Algorithms~\ref{alg:step} and~\ref{alg:task} give the two instantiations of
Equation~\ref{eq:experience-activation}. Neither updates a parameter: every
quantity is either precomputed once from the history with frozen encoders or
computed at inference from the current input.

\begin{algorithm}[htbp]
\caption{Step-level activation for personalized execution}
\label{alg:step}
\small
\begin{algorithmic}[1]
\Require instruction $q$; history $\mathcal M_u$; frozen goal encoder $f$,
  screen representation $\phi$, policy $\pi_\theta$; budget $K_\xi$
\State \textbf{Offline:} cache $f(\rho(g_i))$ for every record and
  $\phi(s_{i,j})$ for every recorded screen
\State $\mathcal C_u(q)\gets\operatorname{TopK}_{i\in\mathcal I_u^\xi;\,K_\xi}\;
  f(q)^\top f(\rho(g_i))$
  \Comment{fixed for the whole task}
\State $e_0\gets$ initial reference from the same records
  (Appendix~\ref{app:entry-preference}); $\mathcal H_0\gets\emptyset$
\For{$t=0,1,2,\ldots$}
  \State observe the current screen $s_t$
  \If{$t>0$}
    \State $(i_t^*,j_t^*)\gets\arg\max_{i\in\mathcal C_u(q),\,0\le j<T_i}
      \phi(s_t)^\top\phi(s_{i,j})$
      \Comment{joint search over trajectories and states}
    \State $e_t\gets\Gamma_{\mathrm{step}}\bigl((i_t^*,j_t^*),x_t\bigr)$
      \Comment{serialize $a_{i_t^*,j_t^*}$}
  \EndIf
  \State $\widehat a_t\gets\pi_\theta(q,s_t,\mathcal H_t,e_t)$
    \Comment{the policy may adopt, adapt, or reject $e_t$}
  \If{$\widehat a_t$ terminates the task} \State \textbf{break} \EndIf
  \State execute $\widehat a_t$; $\mathcal H_{t+1}\gets\mathcal H_t\cup\{\widehat a_t\}$
\EndFor
\end{algorithmic}
\end{algorithm}

\begin{algorithm}[htbp]
\caption{Task-level activation for proactive suggestion}
\label{alg:task}
\small
\begin{algorithmic}[1]
\Require query $(c_x,\tau_x)$; history $\mathcal M_u$; frozen intent
  encoder $\psi$; frozen backbone; $\eta,\sigma,n_{\min},n_{\mathrm{ctx}},\delta,\gamma$
\State \textbf{Offline:} cache $z_i=\psi(g_i)$ and the similarity matrix
  $[z_j^\top z_i]_{j,i\in\mathcal I_u}$
\For{each center $j\in\mathcal I_u$}
  \State $\mathcal R_j(x)\gets\{i:\tau_i\ne\tau_x,\ z_j^\top z_i\ge\eta\}$;
    \quad $n_j\gets|\mathcal R_j(x)|$;
    \quad $n_j^{c_x}\gets\sum_{i\in\mathcal R_j(x)}\mathbb I[c_i=c_x]$
  \State $M_j(x)\gets\sum_{i\in\mathcal R_j(x)}\mathbb I[c_i=c_x]\,
    \exp\!\bigl(-d_{24}(\omega_i,\omega_x)^2/2\sigma^2\bigr)$
\EndFor
\State $\mathcal A(x)\gets\{j:n_j\ge n_{\min},\ n_j^{c_x}\ge n_{\mathrm{ctx}},\
  M_j(x)\ge\delta\}$
  \Comment{eligibility by absolute support}
\If{$\mathcal A(x)=\emptyset$} \State \Return \textsc{Abstain} \EndIf
\State $j^*\gets\arg\max_{j\in\mathcal A(x)} M_j(x)/n_j^{\gamma}$
  \Comment{ranking by match per occurrence}
\State $i^\dagger\gets\arg\max_{i\in\mathcal R_{j^*}(x),\,c_i=c_x}
  \sum_{k\in\mathcal R_{j^*}(x)}z_i^\top z_k$
\State \Return suggestion verbalized by the frozen backbone from the intent
  reference $g_{i^\dagger}$
\end{algorithmic}
\end{algorithm}

\paragraph{Cost.} Both procedures are training free, and their offline work
is linear in the history: each historical screen passes once through the
backbone's visual tower and each goal once through the text encoder. Online,
step-level activation adds one visual encoding of $s_t$ and
$\sum_{i\in\mathcal C_u(q)}T_i$ inner products of dimension $B\cdot d$ per
observation, a cost that is independent of the total history size because
$\mathcal C_u(q)$ holds at most $K_\xi=8$ trajectories. The policy is still
called once per step, as in Base, and reads a single short reference rather
than whole records (Section~\ref{sec:ExpActivator-cost-coldstart}). Task-level activation scans the
precomputed similarity matrix of one user, with at most 414 records in this
benchmark, before the backbone verbalizes the activated intent.

\section{Implementation Details}
\label{app:cue-details}

\subsection{Hyperparameters and Inference Settings}
\label{app:hyperparameters}

Table~\ref{tab:hyperparameters} lists every setting used in the reported
runs. All values are fixed before evaluation and shared by the four
backbones unless stated otherwise.

\begin{table}[htbp]
\centering
\small
\caption{Hyperparameters and inference settings.}
\label{tab:hyperparameters}
\begin{tabular}{@{}lll@{}}
\toprule
Component & Setting & Value \\
\midrule
\multirow{4}{*}{Step-level}
 & goal encoder $f$ & paraphrase-multilingual-MiniLM-L12-v2 \\
 & candidate trajectories $K_\xi$ & 8 \\
 & screen layout $B$ & 4 centered quadrants; MAI-UI adds 1 global block \\
 & reference coordinate frame & normalized to $[0,999]$ \\
\midrule
\multirow{4}{*}{Task-level}
 & intent encoder $\psi$ & paraphrase-multilingual-MiniLM-L12-v2 \\
 & routine threshold $\eta$ & 0.94 \\
 & supports $n_{\min}$, $n_{\mathrm{ctx}}$ & 12, 4 \\
 & bandwidth $\sigma$, mass $\delta$, exponent $\gamma$ & 1 hour, 0.25, 1 \\
\midrule
\multirow{5}{*}{Inference}
 & decoding & greedy (temperature 0), seed 0 \\
 & response and context length & 256 and 8,192 tokens \\
 & screenshot budget & 200,704 pixels, one image per request \\
 & precision and engine & BF16, vLLM 0.28.0 \\
 & hardware & one GPU per run \\
\midrule
\multirow{2}{*}{Evaluation}
 & suggestion encoder & Qwen3-Embedding-0.6B \\
 & bootstrap & 20,000 paired draws over users \\
\bottomrule
\end{tabular}
\end{table}

\subsection{Step-level Activation: Application Choice}
\label{app:entry-preference}

The goal encoder $f$ in Section~\ref{sec:step-activation} is
\texttt{paraphrase-multilingual-MiniLM-L12-v2}. Goal preprocessing removes only
the leading application-opening phrase; the remaining instruction is kept as
the retrieval key. The canonical application label $p_i$ is compiled once
from the historical package name and complete instruction using frozen
Qwen3-VL-8B-Instruct. The same frozen MiniLM encoder embeds complete intents
for routine formation. The official suggestion similarity metric uses
Qwen3-Embedding-0.6B. Visual descriptor normalization is
$\nu(v)=v/\max(\lVert v\rVert_2,\epsilon)$ with $\epsilon=10^{-12}$.

For the initial reference $e_0$ in step-level activation, let $\pi_r$ index the $r$-th most similar same-user record according to
$\alpha_i(q)$. The application reference is obtained by
\begin{equation}
    \begin{aligned}
    V(p\mid q,u)&=\sum_{r=1}^{K_a}
        \mathbb I[p_{\pi_r}=p]\,
        \frac{[\alpha_{\pi_r}(q)]_+}{r},
    \qquad
    p^*=\arg\max_p V(p\mid q,u),
    \\
    e_0&=\texttt{open\_app}(p^*).
    \end{aligned}
    \label{eq:app-vote}
\end{equation}
where $[a]_+=\max(a,0)$. Lexical order resolves exact application-vote ties.
The application reference $e_0$ is supplied through the same frozen-policy interface
as the action references selected at later steps. We use $K_a=10$ and $K_\xi=8$,
selected on 3,226 historical application queries and 1,015 actions from
83 pseudo-held-out trajectories, respectively.

\subsection{Step-level Activation: Screen Representation}
\label{app:visual-descriptors}

Each visual block is the normalized mean of a specified set of tokens from
the GUI backbone under evaluation. Candidate blocks include global means
from visual layers 8, 16, 24, and the final layer, as well as means over the
four quadrants of the final spatial token grid. The spatial layout uses the
four quadrant blocks; the pyramid layout adds the final-layer global block.
Intermediate blocks provide additional calibration candidates. Their readout
uses the backbone's native visual features; Qwen3.5 uses unmerged visual
tokens. Centering statistics $\mu_k$ are estimated from the distinct
historical reference images in the calibration retrieval pools and then
reused during evaluation.

For each user with sufficient readable history, the final readable historical
trajectory provides calibration queries and the earlier trajectories provide
reference candidates. All descriptor alternatives are compared on the same
932 within-trajectory actions for each backbone. Calibration evaluates both raw
normalized blocks and centered blocks. Calibration selects four centered spatial blocks for Qwen3-VL, Qwen3.5,
and GUI-Owl, and four centered spatial blocks plus one global block for
MAI-UI. These screen representations
use Eq.~\ref{eq:visual-state-representation}, with $B=4$ for the spatial
layout and $B=5$ for the pyramid layout.

In Qwen3-VL, Qwen3.5, GUI-Owl, and MAI-UI order, historical action-matching
accuracy is 35.8\%, 35.4\%, 35.1\%, and 36.6\%, compared with 29.6\% for
pixel mean squared error (RGB-MSE). Exact McNemar $p$-values are $1.5\times10^{-4}$,
$1.8\times10^{-4}$, $5.1\times10^{-4}$, and $2.1\times10^{-5}$,
respectively. These calibration results determine the descriptor used for
the subsequent policy evaluation.

\subsection{Step-level Activation: Action Reference}
\label{app:evidence-serialization}

The operator $\Gamma_{\mathrm{step}}$ in Equation~\ref{eq:step-activation}
preserves the historical action type and its
structured arguments. For a historical click at pixel coordinates $(x,y)$ on
a screen of size $(W,H)$, the action reference uses
\begin{equation}
    (x^e,y^e)=\left(
        \operatorname{round}\!\left(999\frac{x}{W}\right),
        \operatorname{round}\!\left(999\frac{y}{H}\right)
    \right).
    \label{eq:coordinate-normalization}
\end{equation}
The same convention applies to long clicks. Scroll direction and typed text
retain their structured fields; wait, back, home, and termination preserve
their action types. These fields describe the historical reference. The
policy receives the current screenshot, instruction, action prefix, and
the action reference and predicts the current operation as described in
Section~\ref{sec:step-activation}.

\subsection{Task-level Activation: Intent Reference}
\label{app:routine-selection}

For task-level activation (Section~\ref{sec:task-activation}), circular time
distance in Eq.~\ref{eq:contextual-mass} is
$d_{24}(a,b)=\min(|a-b|,24-|a-b|)$ for hours in $[0,24)$.
The routine threshold is $\eta=0.94$, minimum total support is
$n_{\min}=12$, minimum scenario support is $n_{\mathrm{ctx}}=4$, temporal
bandwidth is $\sigma=1$ hour, minimum weighted support is $\delta=0.25$,
and frequency normalization exponent is $\gamma=1$.
The exclusion $\tau_i\ne\tau_x$ applies to neighborhood membership; each
historical intent remains eligible to serve as a neighborhood center.

After selecting $j^*$, \methodname{} identifies a representative among the members
sharing the query scenario. Its centrality is computed against every member
of the selected semantic routine:
\begin{equation}
    i^\dagger=\arg\max_{\substack{i\in\mathcal R_{j^*}(x)\\c_i=c_x}}
        \frac{1}{n_{j^*}}
        \sum_{k\in\mathcal R_{j^*}(x)} z_i^\top z_k,
    \qquad e_u^{(\mathrm{task})}(x)=g_{i^\dagger}.
    \label{eq:routine-medoid}
\end{equation}
Source index resolves centrality ties within numerical tolerance
$10^{-7}$. The task-level activation ablations score this historical intent directly,
before suggestion generation. In the backbone comparison, the selected intent is supplied as
an intent reference and the frozen backbone verbalizes the suggestion. Both
evaluations use the same task-level activation rules.

\subsection{Prompt Templates}
\label{app:prompts}

Step-level activation reaches the policy as one structured field appended to
the backbone's native execution prompt; the system prompt, action space, and
output format are those of Base. The user turn reads:

\begin{Verbatim}[frame=single,fontsize=\scriptsize,framesep=2mm]
## User Profile
{user_profile}
## Task Progress
{history_actions}
## User's Query
{instruction}
## Grounded Historical Evidence
{"evidence_type": "action_reference",
 "retrieved_action": {"action": "click", "coordinate": [183, 599]},
 "historical_coordinate_frame": "0-999 normalized",
 "retrieval_basis": "same-user residual-goal trajectory retrieval and
                     visual state matching"}

The evidence is a candidate next action retrieved from personal history; its
retrieval_basis field states whether it comes from a goal-level vote or a
goal-and-screen match. Use it to guide the current decision, but do not copy it mechanically.
Inspect the current screenshot and task progress, then adopt, adapt, or
reject the candidate as needed. For a click or long press, locate the
target on the current screenshot and produce current coordinates.
\end{Verbatim}

For task-level activation, the activated routine is passed to the prompt
below, which HIM-Agent uses for its own routines. The routine is serialized
as JSON with the fields shown; when \methodname{} abstains, the field is an
empty list.

\begin{Verbatim}[frame=single,fontsize=\scriptsize,framesep=2mm]
You are skilled at analyzing user history. You are given summarized user
daily routines. Determine whether the current user state requires a
proactive suggestion.
## Note
- Each summarized routine describes a frequent behavior with intent,
  time/scenario distribution, and frequency.
- If neither the time nor the scenario strongly matches a summarized
  routine, output False.
- If a recommendation is needed, output a suitable user instruction for
  the GUI agent to execute.
- Express the user's intent unambiguously in one Chinese sentence without
  adding extra description.
- Do not output explanations, time, or scenario information. Output only
  the user instruction or False.
## Input
User_profile: {profile}   Time: {time}   Scenario: {scenario}
Summarized_routine: [{"routine_intent": g, "historical_state": {time, scenario},
  "occurrence_count": n_j, "matched_scenario_count": n_j^c,
  "temporal_state_mass": M_j, "normalized_state_likelihood": M_j / n_j}]
The user's intent:
\end{Verbatim}

\section{Benchmark and Evaluation Protocol}
\label{app:evaluation}

\subsection{Dataset Statistics}
\label{app:dataset-stats}

Table~\ref{tab:dataset-stats} summarizes the evaluation data. Histories vary
by an order of magnitude across users, and most target steps are taken on an
interface that the instruction does not describe: clicks and scrolls alone
account for 59\% of all steps.

\begin{table}[htbp]
\centering
\small
\caption{Statistics of the history and evaluation splits. Action shares are
over the 7,915 execution steps.}
\label{tab:dataset-stats}
\begin{tabular}{@{}lr@{\hspace{2.2em}}lr@{}}
\toprule
\multicolumn{2}{@{}l}{History and execution} & \multicolumn{2}{l@{}}{Action type share (\%)} \\
\midrule
Users & 82 & click & 41.1 \\
History records & 15,966 & scroll & 17.9 \\
Records per user (median) & 196.5 & wait & 13.6 \\
Records per user (min, max) & 39, 414 & open app & 9.8 \\
Execution episodes & 775 & finish & 9.8 \\
Execution steps & 7,915 & type & 5.8 \\
Steps per episode (mean, median) & 10.2, 8 & back, home, recent & 1.8 \\
Steps per episode (max) & 85 & long click & 0.3 \\
Proactive states (positive, negative) & 215, 100 & & \\
\bottomrule
\end{tabular}
\end{table}

\subsection{Data and Evaluation Protocol}
\label{app:cue-protocol}

The chronological first $80\%$ of each user's records form the historical
pool. Every target execution episode is excluded from execution retrieval.
Proactive queries contain user identity, time, and scenario; records with
the exact query timestamp are excluded from routine membership as in
Section~\ref{sec:task-activation}. Historical calibration fixes retrieval
neighborhood sizes, descriptor layouts, and centering statistics before
policy evaluation. The semantic encoders, visual towers, application-label
compiler, and GUI policies remain frozen.

$\mathcal H_t$ in Section~\ref{sec:step-activation} is the benchmark's preceding action sequence. Reference selection uses the current instruction and screen, while the
current target action is reserved for scoring. Appendix~\ref{app:leap-ablations}
details the step-level activation experiments on selecting versus appending
and on reference updates, followed by the task-level activation experiments
on time, scenario, and frequency normalization.

\subsection{Metric Definitions}
\label{app:metrics}

\paragraph{Execution.} For a predicted action $\widehat a_t$ and the recorded
action $a_t$, type accuracy counts $\widehat a_t$ as correct when the action
types agree. Step success additionally requires the arguments to agree: a
click or long click must fall within a normalized distance of $0.14$ of the
target on a $1000\times1000$ grid, a scroll must share the direction, typed
text must reach a character similarity of $0.7$, and an application opening
must name the target application; back, home, wait, and finish match on type
alone. SSR averages step success over all steps. The benchmark's cumulative
success rate weights the steps of an episode of length $L$ by
\begin{equation}
    \mathrm{CSR}=\frac{\sum_e\sum_{k<L_e}w_k^{(L_e)}\,m_{e,k}}
                      {\sum_e\sum_{k<L_e}w_k^{(L_e)}},
    \qquad
    w_k^{(L)}=L\,\frac{e^{-0.6k}}{\sum_{k'<L}e^{-0.6k'}},
    \label{eq:csr}
\end{equation}
where $m_{e,k}\in\{0,1\}$ is step success at position $k$ of episode $e$. CSR
therefore emphasizes the early decisions that determine whether a
personalized episode starts on the right path.

\paragraph{Proactive suggestion.} A method emits a suggestion or abstains on
each state. With true and false positives and negatives counted over the 315
states, precision is $P=\mathrm{TP}/(\mathrm{TP}+\mathrm{FP})$, recall is
$R=\mathrm{TP}/(\mathrm{TP}+\mathrm{FN})$, the false-alarm rate is
$\mathrm{FA}=\mathrm{FP}/(\mathrm{FP}+\mathrm{TN})$, and
$\mathrm{BA}=\tfrac12(R+1-\mathrm{FA})$. The Matthews correlation coefficient is
\begin{equation}
    \mathrm{MCC}=
    \frac{\mathrm{TP}\cdot\mathrm{TN}-\mathrm{FP}\cdot\mathrm{FN}}
    {\sqrt{(\mathrm{TP}+\mathrm{FP})(\mathrm{TP}+\mathrm{FN})
           (\mathrm{TN}+\mathrm{FP})(\mathrm{TN}+\mathrm{FN})}},
    \label{eq:mcc}
\end{equation}
which is zero for a rule that suggests on every state and, unlike F1, is not
inflated by the 68\% share of positive states. The semantic
score follows the benchmark: for every suggestion emitted on a positive state
it averages the normalized edit similarity and the Qwen3-Embedding-0.6B
cosine similarity between the suggestion and the recorded intent, and Sem.\
reports the mean over these suggestions.

\paragraph{Statistical comparison.} Every comparison is paired, since all
methods are evaluated on the same steps or states. We resample users with
replacement, keep all of a resampled user's steps or states together, and
recompute the metric difference on each of 20,000 draws; a difference is
called significant when its 95\% percentile interval excludes zero.

\subsection{Baseline Construction}
\label{app:baselines}

All baselines read the same chronological history as \methodname{} and are
inserted through the native prompt of each backbone. \textbf{Recent} supplies
the latest three records with their intent, application, and actions.
\textbf{Retrieve} supplies the three same-user records whose intents are most
similar to the instruction under Qwen3-Embedding-0.6B. \textbf{HIM-Agent}
uses the released history filter, intent clustering with a minimum cluster
size of three, and top-three retrieval with the released record format.
\textbf{LLM-UM} supplies an offline per-user profile of frequent behaviors
with example actions; since the original profiles are not released, we
generate them deterministically with a local Qwen3.5-9B from each user's most
recent 150 history records. For proactive suggestion, HIM-Agent and
\methodname{} share one prompt that presents summarized routines, while
Recent, Retrieve, and LLM-UM share one that presents raw history records.
Every context was fixed for all states and episodes before inference.

\subsection{Complete Proactive Metrics}
\label{app:proactive-complete}
Table~\ref{tab:proactive-complete} lists the benchmark's recall, false-alarm
rate, and F1 alongside precision and the suggestion score. Because 68\% of the
315 states are positive, a rule that suggests on every state attains 81.1 F1
with no ability to tell states apart, which is why the main table summarizes
the activation decision with balanced accuracy and the Matthews correlation
coefficient.

\begin{table}[htbp]
\centering
\tiny
\caption{Complete proactive metrics on 215 positive and 100 negative states.
All values are percentages; lower is better for FA.}
\label{tab:proactive-complete}
\resizebox{\linewidth}{!}{%
\begin{tabular}{llrrrrr}
\toprule
Backbone & Method & Sem. & P & R & FA & F1 \\
\midrule
\multirow{5}{*}{Qwen3-VL-8B}
 & Recent & 48.6 & 68.4 & 98.6 & 98.0 & 80.8 \\
 & Retrieve & 48.4 & 67.8 & 95.8 & 98.0 & 79.4 \\
 & LLM-UM & 29.1 & 68.9 & 99.1 & 96.0 & 81.3 \\
 & HIM-Agent & 50.8 & 72.7 & 87.9 & 71.0 & 79.6 \\
 & \methodname{} & 52.1 & 82.6 & 79.5 & 36.0 & 81.0 \\
\midrule
\multirow{5}{*}{MAI-UI-8B}
 & Recent & 44.5 & 68.3 & 100.0 & 100.0 & 81.1 \\
 & Retrieve & 44.3 & 68.3 & 100.0 & 100.0 & 81.1 \\
 & LLM-UM & 26.7 & 68.3 & 100.0 & 100.0 & 81.1 \\
 & HIM-Agent & 50.5 & 70.6 & 91.6 & 82.0 & 79.8 \\
 & \methodname{} & 49.6 & 82.7 & 82.3 & 37.0 & 82.5 \\
\midrule
\multirow{5}{*}{Qwen3.5-9B}
 & Recent & 47.3 & 69.5 & 95.3 & 90.0 & 80.4 \\
 & Retrieve & 52.1 & 68.2 & 34.0 & 34.0 & 45.3 \\
 & LLM-UM & 21.5 & 61.9 & 6.0 & 8.0 & 11.0 \\
 & HIM-Agent & 48.4 & 82.4 & 41.4 & 19.0 & 55.1 \\
 & \methodname{} & 51.0 & 86.5 & 62.8 & 21.0 & 72.8 \\
\midrule
\multirow{5}{*}{GUI-Owl-1.5-8B}
 & Recent & 46.5 & 69.2 & 96.3 & 92.0 & 80.5 \\
 & Retrieve & 48.2 & 67.8 & 90.2 & 92.0 & 77.4 \\
 & LLM-UM & 26.6 & 67.1 & 70.2 & 74.0 & 68.6 \\
 & HIM-Agent & 51.1 & 72.5 & 86.0 & 70.0 & 78.7 \\
 & \methodname{} & 53.9 & 82.6 & 81.9 & 37.0 & 82.2 \\
\midrule
\multicolumn{2}{l}{Always suggest} & -- & 68.3 & 100.0 & 100.0 & 81.1 \\
\bottomrule
\end{tabular}}
\end{table}

\section{Ablation Protocols and Additional Results}
\label{app:leap-ablations}

\subsection{Step-level Activation: Selecting versus Appending}
\label{app:leap-fixed-history}
\paragraph{Protocol.} We use the existing history split, retrieval scores, and
history-calibrated visual representation without additional parameter tuning.
For each test episode we fix the highest-ranked usable trajectory retrieved
from that user's historical pool. The Whole trajectory condition supplies this
trajectory as a complete record, and the Selected reference condition applies
step-level activation (Section~\ref{sec:step-activation}) to the same
trajectory and supplies one action. Both run the frozen Qwen3-VL-8B policy on
the 7,140 within-trajectory actions with the current screen and the
ground-truth preceding actions visible. A second control, at
$K_\xi=8$, replaces the selected reference with the action immediately
preceding it in the same historical trajectory. Intervals resample the 82
users with replacement 20,000 times, and inference uses greedy decoding with
seed 0, a maximum context length of 8,192 tokens, and up to 256 output tokens.

\paragraph{Results.} Selecting one reference raises policy SSR from 18.47\%
to 23.84\%, a gain of 5.36 [3.74, 7.06] points. At $K_\xi=8$, the preceding
historical action reaches 23.32\% against 26.67\% for the selected reference,
a gain of 3.35 [2.34, 4.46] points. Table~\ref{tab:state-controls} collects
both controls.

\begin{table}[t]
\centering
\small
\caption{Qwen3-VL-8B controls on the 7,140 within-trajectory actions. Whole
trajectory supplies the entire record. Previous
historical action replaces the selected action with the action immediately
before it in the historical trajectory.}
\label{tab:state-controls}
\begin{tabular}{lr}
\toprule
Action reference & Policy SSR \\
\midrule
\multicolumn{2}{l}{\textit{Step-level activation, $K_\xi=1$}} \\
Whole trajectory & 18.47 \\
Selected reference & 23.84 \\
\midrule
\multicolumn{2}{l}{\textit{Step-level activation, $K_\xi=8$}} \\
Previous historical action & 23.32 \\
Selected reference (\methodname{}) & 26.67 \\
\bottomrule
\end{tabular}
\end{table}

\subsection{Step-level Activation: Reference Updates}
\label{app:leap-refresh}
The reference-update experiment uses the frozen Qwen3-VL-8B checkpoint with a
200,704-pixel image budget, 256-token output cap, 8,192-token context limit,
256-request batches, and at most 64 concurrent sequences. The policy receives
the same user profile, ambiguous query, current screen, and action prefix. The reference-update experiment changes only the action reference supplied by step-level activation.
At within-trajectory position $t$, Previous screen uses $s_{\max(1,t-1)}$;
Update every four steps uses $s_{1+4\lfloor(t-1)/4\rfloor}$. Both leave the
reference at the first step unchanged. The retrieved trajectories stay fixed, so the action reference for a stored
observation can be reused exactly.

\begin{table}[htbp]
\centering
\tiny
\caption{Step-level activation with eight fixed historical trajectories. All conditions run the frozen Qwen3-VL-8B policy on the same current inputs and 7,140 within-trajectory actions. Rates are percentages; differences are Current screen minus control in percentage points with 95\% paired user-cluster intervals.}
\label{tab:leap-refresh}
\resizebox{\linewidth}{!}{%
\begin{tabular}{lrl}
\toprule
Action reference & Policy SSR & Difference [95\% CI] \\
\midrule
Previous screen & 22.63 & 4.03 [3.22, 4.84] \\
Update every four steps & 23.17 & 3.50 [2.59, 4.47] \\
Current screen (\methodname{}) & 26.67 & -- \\
\bottomrule
\end{tabular}}
\end{table}

\subsection{Task-level Activation: Time, Scenario, and Frequency}
\label{app:leap-proactive-factors}
Task-level activation ablations use the same frozen MiniLM index over 15,966 historical
intents and expose only user identity, time and scenario to selection. The
full condition reproduces every production suggestion and source record.
No time weighting sets the exponential factor in Equation~\ref{eq:contextual-mass}
to one. No scenario condition removes the scenario indicator, the minimum
same-scenario support requirement, and the same-scenario restriction on the
selected intent. No frequency normalization ranks routines by $M_j(x)$
instead of $M_j(x)/n_j^\gamma$ in Equation~\ref{eq:contextual-score},
while keeping the eligible set unchanged. Activation curves vary the threshold on the maximum support $M_j(x)$,
computed under each condition, among routines meeting the minimum occurrence requirements, with a score of $-1$ for queries without
such a routine. These are ROC diagnostics of the declared score, including
standard endpoints; dots identify the actual selector's fixed-threshold
decisions. The official Semantic scorer uses Qwen3-Embedding-0.6B and fuzzy edit similarity.

\begin{table}[htbp]
\centering\small
\caption{Task-level activation before suggestion generation on 215 positive and 100 negative states. BA is balanced accuracy and FA is false-alarm rate. Semantic is the official score on emitted positive suggestions. All conditions use the same activation threshold. All values are percentages.}
\label{tab:leap-proactive-factors}
\resizebox{\linewidth}{!}{%
\begin{tabular}{lrrrrr}
\toprule
Condition & AUROC & BA & Recall & FA & Semantic \\
\midrule
\methodname{} & 80.37 & 72.66 & 82.33 & 37.00 & 55.35 \\
No time weighting & 74.10 & 69.33 & 84.65 & 46.00 & 50.61 \\
No scenario condition & 79.95 & 64.58 & 91.16 & 62.00 & 52.67 \\
No frequency normalization & 80.37 & 72.66 & 82.33 & 37.00 & 53.24 \\
\bottomrule
\end{tabular}}
\end{table}

At the declared operating point, balanced accuracy falls by 3.34 [0.15, 6.74]
points without time weighting and by 8.08 [2.64, 13.70] points without the
scenario condition, and removing time weighting also lowers AUROC by 6.28
[1.15, 11.96] points. Frequency normalization raises Semantic on emitted
suggestions by 2.11 [0.61, 4.10] points without changing any trigger
decision.

\begin{figure}[htbp]
\centering
\includegraphics[width=0.85\linewidth]{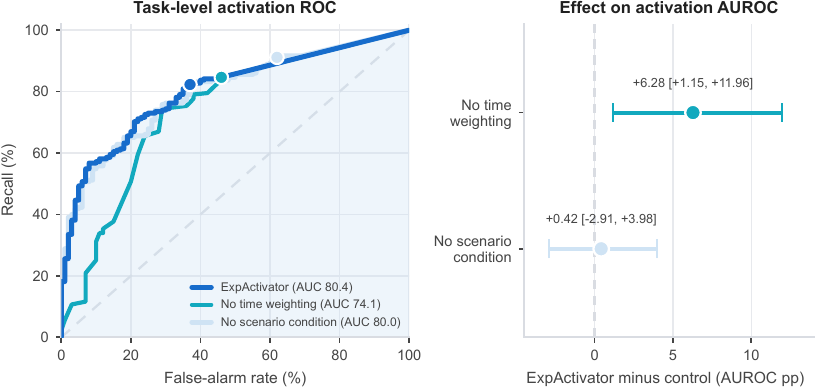}
\caption{Task-level activation ROC curves and paired AUROC differences.
Dots mark the original threshold. Error bars are 95\% user-cluster bootstrap
intervals. No frequency normalization has the same trigger decisions and
curve as \methodname{}. The curves include the standard ROC endpoints described above.}
\label{fig:leap-activation-curves}
\end{figure}


\section{Threshold Sensitivity of Task-level Activation}
\label{app:leaf-proactive-sensitivity}

Task-level activation declares six constants: the routine threshold $\eta$, the
minimum total support $n_{\min}$, the minimum same-scenario support
$n_{\mathrm{ctx}}$, the temporal bandwidth $\sigma$, the activation threshold
$\delta$, and the normalization exponent $\gamma$. This section re-runs the
activation rule of Section~\ref{sec:task-activation} over the sealed routine
index for grids of these constants, so the reported operating point can be read
against the surface it sits on. Only activation is re-run: no suggestion is
generated and no model is called. The reimplementation reproduces the sealed
run exactly at the declared setting (AUROC 80.37, balanced accuracy 72.66,
recall 82.33, false alarms 37.00), which is asserted before any sweep. The
sweeps are diagnostic: every reported result uses the declared constants.

\paragraph{Routine threshold and minimum support.}
Over $\eta\in[0.88,0.98]$ and $n_{\min}\le 12$, activation AUROC stays above
75 in all 40 cells and varies by less than 5.4 points
(Figure~\ref{fig:leaf-proactive-sensitivity}); the declared setting
$(\eta,n_{\min})=(0.94,12)$ reaches 80.37, within 0.33 points of the grid
maximum, so the operating point sits on a broad plateau rather than a sharp
peak. Beyond $n_{\min}=12$ the surface falls away, to
73.24 at $n_{\min}=16$ and 58.95 at $n_{\min}=24$. This is a coverage effect
rather than an instability of the score: requiring that many repetitions leaves
progressively more states with no eligible routine at all, rising from 27.6\%
at $n_{\min}=12$ to 41.9\% at 16 and 73.0\% at 24, and those states receive the
constant abstention score, which removes the discrimination the metric
measures. The usable range of $n_{\min}$ is therefore bounded above by how
often the user's routines actually repeat, not by a tuned threshold.

\begin{figure}[htbp]
\centering
\includegraphics[width=0.85\linewidth]{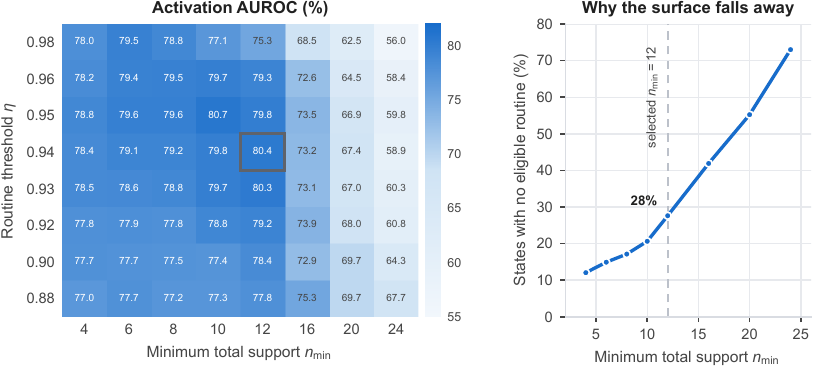}
\caption{Sensitivity of task-level activation. Left: activation AUROC over the
routine threshold $\eta$ and the minimum total support $n_{\min}$, with the
declared setting outlined. Right: the share of the 315 states for which no
routine satisfies the support requirements, at $\eta=0.94$. The AUROC decline
beyond $n_{\min}=12$ tracks this loss of coverage.}
\label{fig:leaf-proactive-sensitivity}
\end{figure}

\paragraph{Bandwidth, scenario support, and activation threshold.}
Table~\ref{tab:leaf-proactive-sensitivity} sweeps the two remaining support
constants one at a time. Temporal bandwidth is flat around its declared value:
AUROC changes by less than 0.6 points for $\sigma\in[0.25,2]$ hours and
declines only once the kernel spans several hours, when time stops
distinguishing routines. Same-scenario support moves AUROC by less than 1.8
points over $n_{\mathrm{ctx}}\in[1,10]$, while larger values steadily lower
the false-alarm rate. The activation threshold $\delta$ cannot change AUROC,
because it selects an operating point on a fixed ranking; over
$\delta\in[0.05,1.0]$ balanced accuracy changes by less than 1.3 points.
Figure~\ref{fig:leap-activation-curves} shows the full
operating-point trace.

\begin{table}[htbp]
\centering\small
\caption{One-at-a-time sensitivity of task-level activation on 215 positive and
100 negative states, with the other constants held at their declared values.
BA is balanced accuracy, FA is the false-alarm rate, and No routine is the
share of states for which no routine meets the support requirements. Declared
settings are marked $^{\dagger}$. All values are percentages except the swept
constant.}
\label{tab:leaf-proactive-sensitivity}
\resizebox{\linewidth}{!}{%
\begin{tabular}{rrrrrr@{\hskip 2em}rrrrrr}
\toprule
\multicolumn{6}{c}{Temporal bandwidth $\sigma$ (hours)}
& \multicolumn{6}{c}{Same-scenario support $n_{\mathrm{ctx}}$} \\
\cmidrule(lr){1-6}\cmidrule(lr){7-12}
$\sigma$ & AUROC & BA & Recall & FA & No routine
& $n_{\mathrm{ctx}}$ & AUROC & BA & Recall & FA & No routine \\
\midrule
0.25 & 79.92 & 73.05 & 72.09 & 26.00 & 27.6 & 1 & 79.82 & 69.59 & 84.19 & 45.00 & 21.6 \\
0.5 & 79.96 & 73.07 & 78.14 & 32.00 & 27.6 & 2 & 79.77 & 69.90 & 82.79 & 43.00 & 24.4 \\
0.75 & 80.20 & 72.73 & 80.47 & 35.00 & 27.6 & 3 & 80.07 & 71.66 & 82.33 & 39.00 & 27.0 \\
1$^{\dagger}$ & 80.37 & 72.66 & 82.33 & 37.00 & 27.6 & 4$^{\dagger}$ & 80.37 & 72.66 & 82.33 & 37.00 & 27.6 \\
1.5 & 80.51 & 71.40 & 82.79 & 40.00 & 27.6 & 5 & 79.74 & 71.50 & 80.00 & 37.00 & 28.9 \\
2 & 80.26 & 70.90 & 82.79 & 41.00 & 27.6 & 6 & 79.62 & 71.50 & 80.00 & 37.00 & 29.2 \\
3 & 79.24 & 70.86 & 83.72 & 42.00 & 27.6 & 8 & 78.94 & 71.57 & 78.14 & 35.00 & 32.4 \\
4 & 78.11 & 69.09 & 84.19 & 46.00 & 27.6 & 10 & 78.59 & 71.44 & 74.88 & 32.00 & 34.9 \\
6 & 76.98 & 69.33 & 84.65 & 46.00 & 27.6 & \multicolumn{6}{c}{} \\
\bottomrule
\end{tabular}}
\end{table}

\paragraph{Frequency normalization.}
The exponent $\gamma$ ranks routines inside the eligible set and cannot move an
activation decision. Setting $\gamma\in\{0,0.5,1.5\}$ leaves every trigger
decision identical to the declared $\gamma=1$ while changing the selected
routine on 100, 41, and 44 of the emitted states, respectively. Its effect is
therefore confined to which intent is verbalized, which
Appendix~\ref{app:leap-proactive-factors} scores with the official suggestion
metric.

\section{Context Cost and History-Size Protocols}
\label{app:leaf-cost}

\subsection{Token accounting}

Every prompt is rebuilt with the evaluator's own message assembly, chat
template, and tokenizer, using the checkpoint revision that produced the run,
and is then tokenized without inference. The screenshot is supplied to the
policy as a separate visual input that is byte-identical across conditions, so
the text-token difference from Base isolates the history each mechanism adds.
The serialized action reference is counted exactly as
it is sent, including a constant field naming the retrieval basis, so the
reported cost of \methodname{} is an overestimate. Base is the same no-history
condition reported in Table~\ref{tab:combined-personalized-proactive}. The
opening and within-trajectory groups partition the evaluation by position in
the episode, which in this benchmark coincides with a partition by kind of
decision: the first recorded action of every episode commits it to an
application, and every later action is taken on an observed interface state.

\begin{figure}[htbp]
\centering
\includegraphics[width=0.85\linewidth]{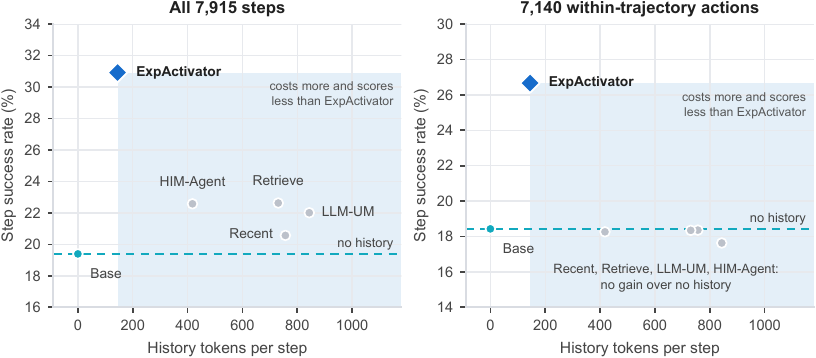}
\caption{Step success against history tokens per step on Qwen3-VL-8B. The
shaded region holds every point that costs more and scores less than
\methodname{}. Left: all steps. Right: actions within the trajectory, where the
four memory baselines stay at the no-history level.}
\label{fig:leaf-cost-pareto}
\end{figure}

\subsection{Within-Trajectory Gains on All Backbones}
\label{app:within-all}

Table~\ref{tab:within-all} repeats the within-trajectory comparison of
Table~\ref{tab:ExpActivator-cost-decomposition} on every backbone. Across the sixteen
pairings of a memory baseline with a backbone, the gain over using no history
never exceeds 2.4 points and is indistinguishable from zero in twelve, while
\methodname{} improves every backbone.

\begin{table}[htbp]
\centering\small
\caption{Within-trajectory step success over the 7,140 actions after each
episode's first decision. Base is the no-history rate in percent; other columns
give the difference from Base in percentage points with paired 95\%
user-cluster bootstrap intervals (20,000 draws).}
\label{tab:within-all}
\resizebox{\linewidth}{!}{%
\begin{tabular}{lrlllll}
\toprule
Backbone & Base & Recent & Retrieve & LLM-UM & HIM-Agent & \methodname{} \\
\midrule
Qwen3-VL-8B & 18.43 & $-0.07$ [$-0.87$, $+0.85$] & $-0.08$ [$-1.16$, $+0.99$] & $-0.80$ [$-1.83$, $+0.12$] & $-0.17$ [$-0.89$, $+0.57$] & $+8.24$ [$+6.51$, $+10.16$] \\
MAI-UI-8B & 17.41 & $+0.77$ [$-0.07$, $+1.63$] & $+2.34$ [$+1.32$, $+3.36$] & $+0.41$ [$-0.39$, $+1.27$] & $+0.98$ [$+0.21$, $+1.75$] & $+5.67$ [$+4.73$, $+6.69$] \\
Qwen3.5-9B & 16.57 & $-0.20$ [$-1.04$, $+0.73$] & $+0.88$ [$-0.27$, $+2.11$] & $-0.10$ [$-1.05$, $+0.86$] & $+0.56$ [$-0.37$, $+1.50$] & $+3.74$ [$+2.79$, $+4.76$] \\
GUI-Owl-1.5-8B & 22.30 & $+0.48$ [$-0.59$, $+1.58$] & $+1.72$ [$+0.75$, $+2.72$] & $+0.63$ [$-0.17$, $+1.42$] & $+1.92$ [$+1.24$, $+2.64$] & $+2.38$ [$+0.53$, $+4.41$] \\
\bottomrule
\end{tabular}}
\end{table}

\subsection{History-size stratification}
\label{app:history-size}
\methodname{} exceeds the strongest baseline within every quartile of history
size by at least 6.4 points, with all four intervals excluding zero
(Figure~\ref{fig:leaf-history-size}); within the trajectory, the strongest
comparator is the no-history condition itself in three of the four groups.
Users in the smallest quartile, with at most 111 records, gain 8.31
[5.32, 12.50] points, so step-level activation is usable from a small personal
history.

\begin{figure}[htbp]
\centering
\includegraphics[width=\linewidth]{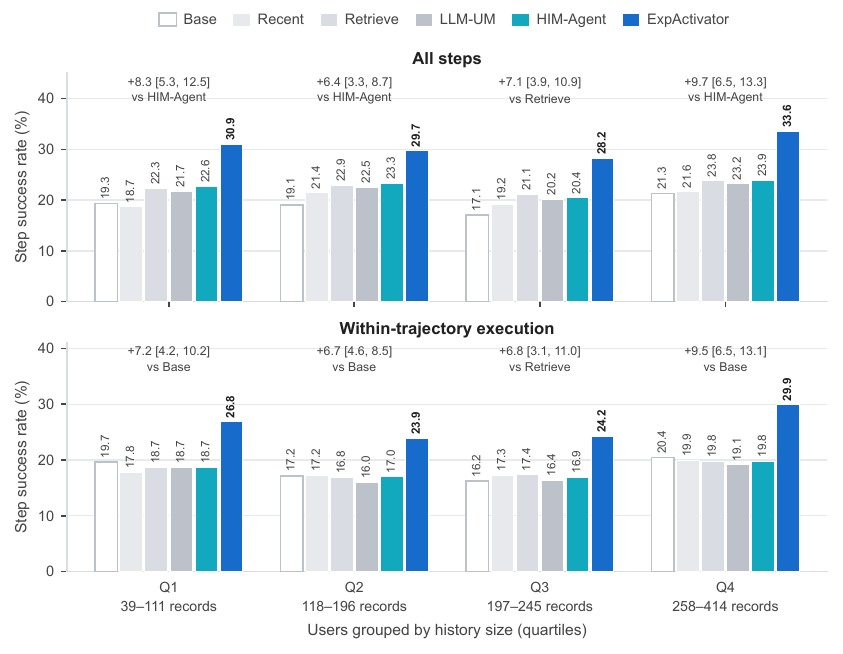}
\caption{Personalized execution by history size on Qwen3-VL-8B. Users are
grouped into quartiles by the number of records in their memory split, and all
six conditions are scored on the same steps inside each quartile. Top: all
steps. Bottom: within-trajectory execution, where the four memory baselines
stay at the no-history level in every quartile. Blue text gives \methodname{}
minus the strongest baseline in that group, in percentage points with paired
95\% user-cluster bootstrap intervals; the strongest baseline is selected per
group and is named, so the comparator changes across groups.}
\label{fig:leaf-history-size}
\end{figure}

User history size is the number of records in that user's chronological
first-80\% memory split, the same pool all methods retrieve from. The 82
evaluated users are divided into quartiles by this count; quartile boundaries
come from the history alone and use no evaluation output. Within each quartile
every condition is scored on the same steps, and intervals resample whole users
with replacement 20,000 times, as elsewhere in this paper. The strongest
baseline is chosen separately inside each quartile, so the comparator is not
fixed across buckets and the reported gain is the smallest available against
any baseline there. Quartiles contain unequal step counts because episode length and the
number of test episodes vary across users.

\begin{table}[htbp]
\centering\small
\caption{Personalized execution by user history size on Qwen3-VL-8B. Users are
grouped into quartiles by records in their memory split. Rates are step success
percentages on the steps belonging to each group. Gain is \methodname{} minus
the strongest baseline within that group, in percentage points with paired 95\%
user-cluster bootstrap intervals.}
\label{tab:leaf-history-size}
\resizebox{\linewidth}{!}{%
\begin{tabular}{lrrrrrrrrrll}
\toprule
Group & Users & Records & Steps & Base & Recent & Retrieve & LLM-UM
& HIM-Agent & \methodname{} & Strongest & Gain [95\% CI] \\
\midrule
\multicolumn{12}{l}{\textit{All steps}} \\
Q1 & 21 & 39--111 & 782 & 19.31 & 18.67 & 22.25 & 21.74 & 22.63 & 30.95 & HIM-Agent & $+8.31$ [$+5.32$, $+12.50$] \\
Q2 & 20 & 118--196 & 1535 & 19.09 & 21.43 & 22.87 & 22.54 & 23.26 & 29.71 & HIM-Agent & $+6.45$ [$+3.29$, $+8.74$] \\
Q3 & 20 & 197--245 & 2424 & 17.08 & 19.22 & 21.08 & 20.21 & 20.42 & 28.22 & Retrieve & $+7.14$ [$+3.90$, $+10.86$] \\
Q4 & 21 & 258--414 & 3174 & 21.33 & 21.64 & 23.79 & 23.19 & 23.88 & 33.55 & HIM-Agent & $+9.67$ [$+6.48$, $+13.25$] \\
All & 82 & 39--414 & 7915 & 19.39 & 20.57 & 22.63 & 22.01 & 22.58 & 30.92 & Retrieve & $+8.29$ [$+6.37$, $+10.40$] \\
\midrule
\multicolumn{12}{l}{\textit{Within-trajectory execution}} \\
Q1 & 21 & 39--111 & 697 & 19.66 & 17.79 & 18.65 & 18.65 & 18.65 & 26.83 & Base & $+7.17$ [$+4.24$, $+10.25$] \\
Q2 & 20 & 118--196 & 1344 & 17.19 & 17.19 & 16.82 & 16.00 & 16.96 & 23.88 & Base & $+6.70$ [$+4.59$, $+8.46$] \\
Q3 & 20 & 197--245 & 2242 & 16.24 & 17.31 & 17.35 & 16.37 & 16.95 & 24.17 & Retrieve & $+6.82$ [$+3.07$, $+10.96$] \\
Q4 & 21 & 258--414 & 2857 & 20.44 & 19.88 & 19.78 & 19.15 & 19.81 & 29.89 & Base & $+9.45$ [$+6.47$, $+13.05$] \\
All & 82 & 39--414 & 7140 & 18.43 & 18.36 & 18.35 & 17.63 & 18.26 & 26.67 & Base & $+8.24$ [$+6.51$, $+10.16$] \\
\bottomrule
\end{tabular}}
\end{table}

Within the trajectory the strongest comparator is the no-history condition in
three of the four quartiles, consistent with the aggregate decomposition in
Table~\ref{tab:ExpActivator-cost-decomposition}. Both analyses are
reproduced by \texttt{analyze\_\allowbreak ExpActivator\_\allowbreak context\_\allowbreak cost.py} and
\texttt{analyze\_\allowbreak ExpActivator\_\allowbreak history\_\allowbreak size.py} from the sealed run files; neither
performs additional model inference.

\section{Case Study}
\label{app:case-study}

Figure~\ref{fig:case-study} traces step-level activation on a real task. The
references for steps 2 and 6 come from an April 9 trip that leaves the same
residence for AS3, because those screens match the current ones. Once the
route screens for Kent Ridge MRT appear, the reference moves to an April 11
trip to that destination, and the frozen policy selects route A1 and ends the
task.

\begin{figure}[ht]
\centering
\includegraphics[width=\linewidth]{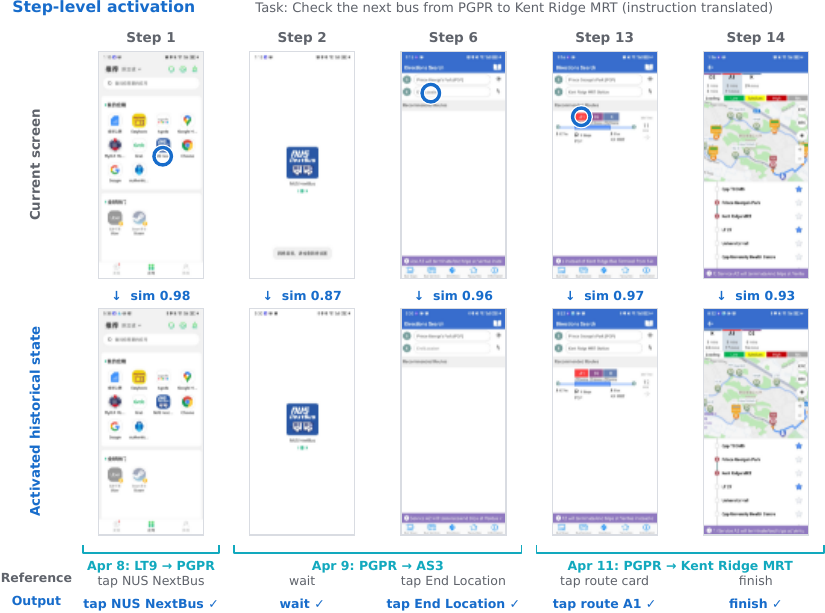}
\caption{\textcolor{StepBlue}{Step-level activation} on five steps of one NUS
NextBus task. Each current screen activates the most similar recorded state
among eight task-relevant trajectories, so the reference moves from one past
trip to another as the screen changes; circles mark the frozen policy's taps.
The task instruction is translated from Chinese.}
\label{fig:case-study}
\end{figure}

\subsection{Task-level Activation for More Users}
\label{app:task-cases}

Figures~\ref{fig:task-more-a} and~\ref{fig:task-more-b} repeat
Figure~\ref{fig:task-case} (right) for five more users, with every routine that
is activated at some hour shown. In every case the activated
routine follows the time of day, and for the commuter it also follows the
scenario: two minutes apart, the same user is offered the e-bus card on the
street and a study task once on board. Every marked state activates the routine
the user performed, and every user has hours in which no routine qualifies and
\methodname{} abstains.

\begin{figure}[htbp]
\centering
\includegraphics[width=\linewidth]{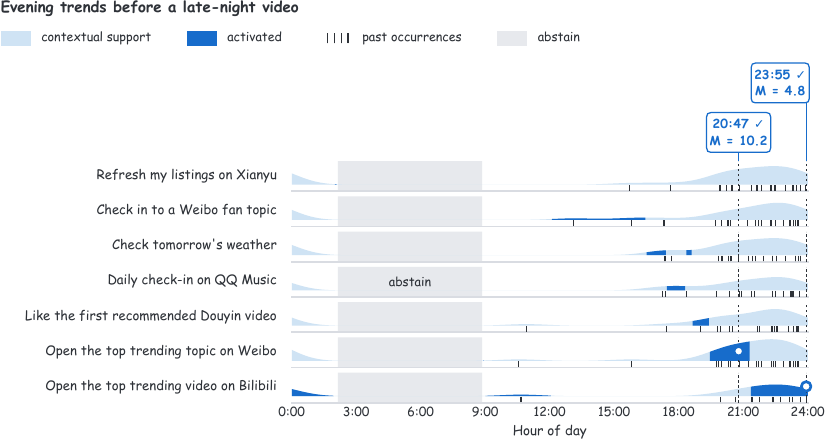}\\[6pt]
\includegraphics[width=\linewidth]{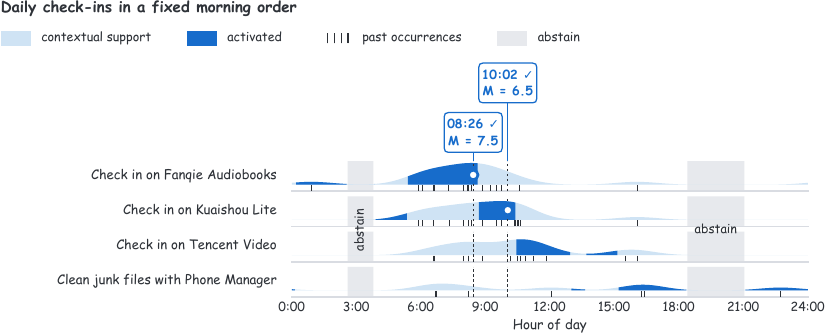}
\caption{Task-level activation for two more users, in the format of
Figure~\ref{fig:task-case} (right). Intents are translated from Chinese.}
\label{fig:task-more-b}
\end{figure}

\begin{figure}[htbp]
\centering
\includegraphics[width=\linewidth]{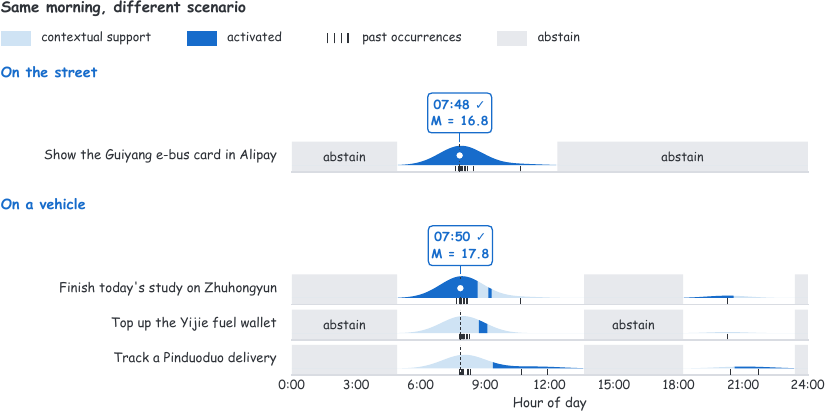}\\[6pt]
\includegraphics[width=\linewidth]{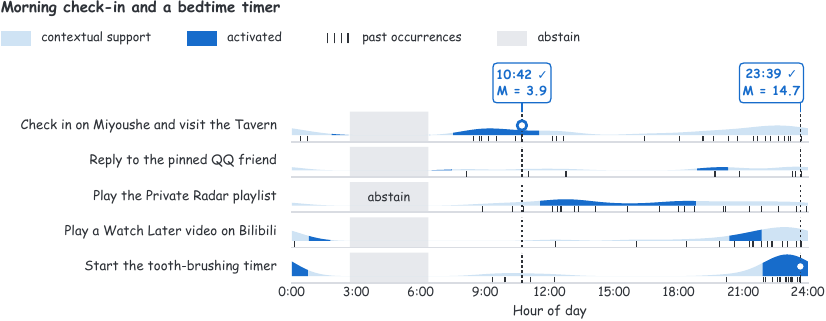}\\[6pt]
\includegraphics[width=\linewidth]{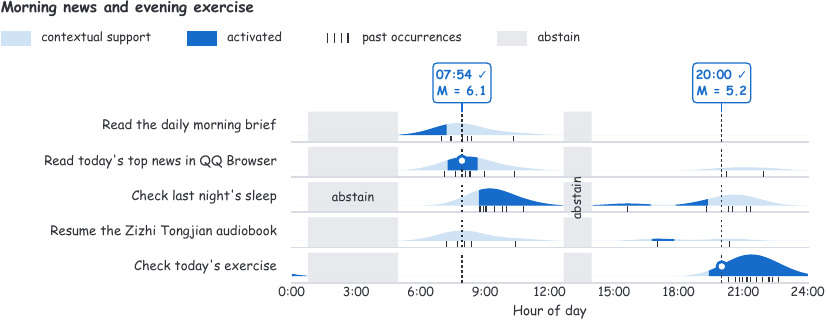}
\caption{Task-level activation for three more users, in the format of
Figure~\ref{fig:task-case} (right). Top: one commuter on the street and on a vehicle on
the same morning. Intents are translated from Chinese.}
\label{fig:task-more-a}
\end{figure}

\end{document}